\documentclass[acmtog, authorversion]{acmart}
\acmSubmissionID{153}

\usepackage{booktabs} % For formal tables
\usepackage{wrapfig}
\usepackage{mathtools}

\usepackage[ruled]{algorithm2e} % For algorithms

\SetAlFnt{\small}
\SetAlCapFnt{\small}
\SetAlCapNameFnt{\small}
\SetAlCapHSkip{0pt}

\acmJournal{TOG}
\acmYear{2020}\acmVolume{39}\acmNumber{4}\acmArticle{129}\acmMonth{7} \acmDOI{10.1145/3386569.3392380}

\setcopyright{acmlicensed}
\begin{document}

%%%%%% use negative v-space
\def\negativevspace{}

%%%%%% formal version
%\newcommand{\TODO}[1]{{\color{red}{[TODO: #1]}}}
%\newcommand{\rz}[1]{{\color{blue}#1}}
%\newcommand{\xh}[1]{{\color[rgb]{0.1,0.46,0.9}#1}}
%\newcommand{\ed}[1]{{\color[rgb]{0.5,0.0,0.5}#1}}
%\newcommand{\phil}[1]{{\color[rgb]{0.3,0.7,0.3}#1}}

\newcommand{\todo}[1]{{\color{red}{[TODO: #1]}}}
\newcommand{\rz}[1]{{\color{black}#1}}
\newcommand{\xh}[1]{{\color{black}#1}}
\newcommand{\ed}[1]{{\color{black}#1}}
\newcommand{\phil}[1]{{\color{black}#1}}

\newcommand{\final}[1]{{\color{black}#1}} %\phil{avoid saturated colors}
\newcommand{\finaltwo}[1]{{\color{black}#1}} %\phil{avoid saturated colors}

% Title portion
\title{TilinGNN: Learning to Tile with Self-Supervised Graph Neural Network}

%Convolutional

% DO NOT ENTER AUTHOR INFORMATION FOR ANONYMOUS TECHNICAL PAPER SUBMISSIONS TO SIGGRAPH 2019!
\author{Hao Xu}
\authornote{Both authors contributed equally to the paper.}
\author{Ka-Hei Hui}
\authornotemark[1]
\author{Chi-Wing Fu}
\affiliation{%
	\institution{The Chinese University of Hong Kong}}
\author{Hao Zhang}
\affiliation{%
	\institution{Simon Fraser University}}
\renewcommand\shortauthors{Xu et al.}

%% This is the ``teaser'' command, which puts an figure, centered, below
%%% the title and author information, and above the body of the content.

\begin{teaserfigure}
  \centerline{\includegraphics[width=0.99\textwidth]{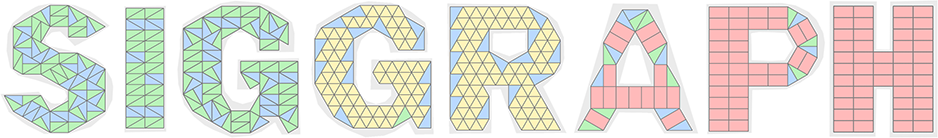}}
  \ifdefined\negativevspace
  \vspace*{-2.5mm}
  \fi
  \caption{Our self-supervised neural network, TilinGNN, produces tiling results in time roughly {\em linear\/} to the number of candidate tile locations, significantly outperforming
  traditional combinatorial search methods. The average runtime of our network for tiling a character is only 25.71s.
  %The letter shapes are of the font type ``Franklin Gothic Heavy'' and 
  The character shapes to be tiled are shown in grey and different types of tiles are displayed using different colors (note that mirror reflections count as different tile types).}
%s
%Richard's conflict	
%	\caption{Our self-supervised neural network, TilinGNN, produces tiling results in time roughly {\em linear\/} to the number of candidate tile locations, significantly outperforming
%	traditional combinatorial search methods. The average runtime of our network for tiling a character is only 25.71s. 
%The letter shapes are of the font type ``Franklin Gothic Heavy'' and 
%	The input character shapes are shown in grey and different types of tiles are displayed in different colors (note that mirror reflections count as different tile types).}
  %\xh{FYI, The font of the input characters is called ``Franklin Gothic Heavy''.}}
  %\rz{TBD: I would try to title SIGGRAPH. Make sure that we mention timing here, to contrast what we said in the intro about how slow the standard solvers are! \xh{The average running time for each character is 44.24s.}} \newline}
  %
  %\caption{\rz{TBD: I would try to title SIGGRAPH. Make sure that we mention timing here, to contrast what we said in the intro about how slow the standard solvers are! \xh{The average running time for each character is 44.24s.}} \newline}
  % avg. time for first three characters: 15.24; middle two: 14.18; last three: 43.87
  \label{fig:teaser}
\end{teaserfigure}

\begin{abstract}
We introduce the first {\em neural optimization\/} framework to solve a classical instance of the {\em tiling\/} problem.
Namely, we seek a {\em non-periodic\/} tiling of an arbitrary 2D shape using {\em one or more} types of tiles---the tiles {\em maximally\/} fill the shape's interior without overlaps or holes. 
To start, we reformulate tiling as a graph problem by modeling candidate tile locations in the target shape as graph nodes and connectivity between tile locations as edges.
Further, we build a {\em graph convolutional neural network\/}, coined TilinGNN, to progressively propagate and aggregate features over graph edges and predict tile placements.
TilinGNN is trained by maximizing the tiling coverage on target shapes, while avoiding overlaps and holes between the tiles.
Importantly, our network is {\em self-supervised\/}, as we articulate these criteria as loss terms defined on the network outputs, without the need of ground-truth tiling solutions.
After training, the runtime of TilinGNN is roughly {\em linear\/} to the number of candidate tile locations, significantly outperforming traditional combinatorial search.
We conducted various experiments on a variety of shapes to showcase the speed and versatility of TilinGNN. We also present comparisons to alternative methods and manual solutions, robustness analysis, and ablation studies to demonstrate the quality of our approach.
Code is available at \url{https://github.com/xuhaocuhk/TilinGNN/}
\end{abstract}


%
% End generated code
%


\maketitle

\section{Introduction}
\label{sec:intro}

Many geometry processing tasks in computer graphics require solving discrete and combinatorial optimization problems, 
e.g., decomposition, packing, set cover, and assignment. Conventional approaches resort to approximation algorithms 
with guaranteed bounds or heuristic schemes exhibiting favorable average performance. With the rapid adoption 
of machine learning techniques in all facets of visual computing, an intriguing question is whether difficult combinatorial 
problems involving geometric primitives can be effectively and efficiently solved using a {\em machine learning\/} approach.

\begin{figure}[b]
	\centering
	\includegraphics[width=0.99\linewidth]{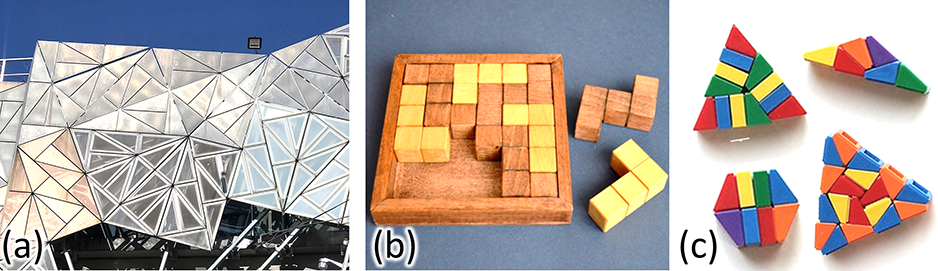}
	\ifdefined\negativevspace
	\vspace*{-1mm}
	\fi
	\caption{Applications of tilings:
		(a) Federation Square in Melbourne;
		(b) a puzzle called jags and hooks \final{designed by Erhan Cubukcuoglu};
		and (c) cheese slope mosaic using LEGO bricks \final{from Katie Walker's Flickr page}.
		%\phil{for (b), if our tiling method cannot work with this tile set, choose a toy with polyominoes}
	}
	%and (c) the Chinese Tangram.
	%\phil{find some other examples to replace (b) and (c)? since our current method cannot really do (c) due to too many tiles and cannot do (b)?  Find real-world examples like (a) if possible}}
	\label{fig:examples}
\end{figure}

In this paper, we explore a learning-based approach to solve a combinatorial geometric optimization problem, {\em tiling\/},
which has drawn interests from the computer graphics community in different contexts~\cite{kaplan-2009}, e.g., 
sampling~\cite{kopf-2006-recursive,ostromoukhov-2007-polyominoes}, texture generation~\cite{cohen-2003-WangTile}, 
architectural construction~\cite{fu-2010-Kset,eigensatz-2010-paneling,singh-2010-triangle}, and puzzle design~\cite{Duncan-2017-ApproximateDissection}.
In general, tiling refers to the partition of a domain into regions, the tiles, of one or more types. So far, most works on
tiling have stayed in the 2D domain; see Figure~\ref{fig:examples} for some 
%well-known 
%\phil{well-known may not be a good word, given these examples?}
%\phil{typical} examples and applications.
typical examples and applications.

%%%%%%%%%%%%%%%%%%%%%%%%%%%%%%%%%%%%%%%%%%%%%%%%%%%%
%%%%%%%%%%%%%%%%%%%%%%%%%%%%%%%%%%%%%%%%%%%%%%%%%%%%
\begin{figure*}[t]
	\centering
	\includegraphics[width=0.99\linewidth]{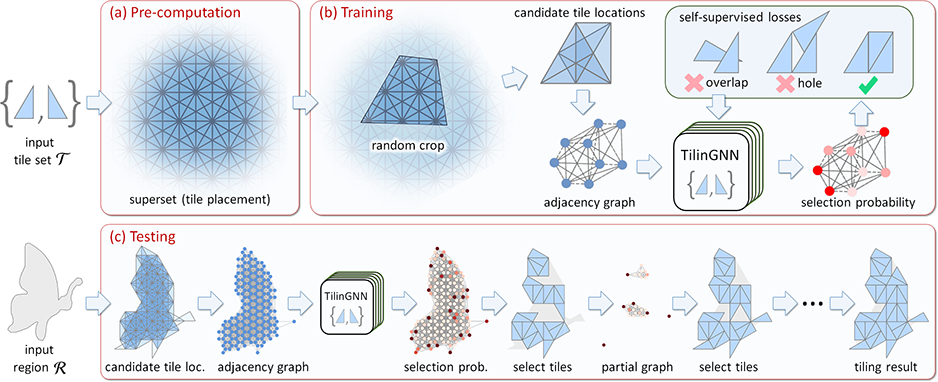}
	\vspace*{-1mm}
	%\ifdefined\negativevspace
	%\vspace*{-1mm}
	%\fi
	\caption{Overview of our ``learn-to-tile" approach:
		(a) we precompute a ``superset'' of candidate tile locations for each given tile set;
		(b) we train our graph neural network TilinGNN to learn to tile for the given tile set by generating random 2D shapes to \final{locate full tile placements inside both the superset and random shape} and then by training TilinGNN to learn to predict locations to place tiles in a self-supervised manner; and
		(c) given a region to tile, we first locate candidate tile locations within the region, then make use of the trained network model to progressively select tile locations and generate the tiling result.
		%\phil{To XH: for the example shown in the ``Testing'' part, did you run your program to see if the first two tiles to be selected are really the two red tiles above? They didn't look to be the best choice in the initial step}
		%\xh{This is the real output of the program. We ever force the tile selection start from corner or boundary, but it is not as good as let a network to decide the order.}
%\phil{but the problem with this result is that it may violate what we said... shall we still say that the network predicts good tile placement locations? or find another example? or? Any idea?}
	}
	%
	%\caption{Overview of our framework. 
	%	Starting from the tile set (a), we first find all possible tile placements around a seed tile (b). 
	%	Then we generate training data of random polygons, and locate the tiles placements inside it (d) by cropping.
	%	The tile placements are then converted an adjacency graph (e), where each node represents a tile placement.
	%	The data are sent as the input of our network (f), which output the probability (g) of selecting each node. 
	%	We evaluate the output using self-supervised losses (h) and update the weight of the network accordingly to train it.
	%	After the training, the test process starts from a new input tiling region (i), which will also be converted into an adjacency graph as the %input of the trained model.
	%	We use a simple procedure to iteratively select tiles (j), guided by the output probability map.
	%	The procedure may call the network a few more times by input the adjacency graph of the rest tiles (k), before arriving at the final tiling %result (l). 
	%	%    User may revise the input according to the generated tiling solution (Figure~\ref{fig:overview} (m)).
	%}
	%
	\label{fig:overview}
	\vspace*{-1mm}
\end{figure*}

As a first attempt, we focus on {\em non-periodic\/}
%{\em aperiodic\/} 
%\phil{Maybe non-periodic? \url{https://en.wikipedia.org/wiki/Aperiodic_tiling}}
%
%
tiling\footnote{\finaltwo{Non-periodicity means that no finite shifts of a tiling can reproduce it. A better known special case of such tilings are {\em aperiodic\/} tilings, e.g., Penrose tilings. Aperiodicity has the additional requirement that the tiling cannot contain arbitrarily large periodic patches, %~\cite{wiki:aperiodic}, 
which is not necessarily respected by our method (see Figure~\ref{fig:teaser}).}} 
% tile set is {\em not\/} aperiodic; it can generate periodic/non-periodic tilings (see Figure~\ref{fig:teaser}).}
of an
%a given 2D region of 
arbitrary 2D shape using {\em one or more} types 
of tiles.
Specifically, we seek a tiling 
%which
\phil{that} 
{\em maximally\/} fills the shape's {\em interior\/}
%of the input region, i.e., 
without
overlaps, holes, or tiles exceeding the shape boundary.
Even such an elementary version of the tiling problem is hard---it is known that whether a finite 2D region 
can be tiled with a given set of tiles is NP-complete, even when the tile set has
%\phil{just to shorten}
%consists of 
only one type such as the 
tromino~\cite{moore-2001}. 
\final{Existing open-source libraries such as COIN-OR/BC~\shortcite{coin-bc} (for branch \& cut mixed integer programming) and commercial combinatorial optimization tools such as Gurobi~\shortcite{gurobi} (specialized integer programming solver) are alternatives to solving our tiling problem.}
%Existing combinatorial search tools that are available from \final{commercial software
%libraries such as Gurobi~\shortcite{gurobi} and open-source libraries, such as} COIN-OR/BC~\shortcite{coin-bc} (for branch-n-cut mixed integer programming) can be applied to our tiling problem.
They are, however, very time-consuming, e.g.,~\final{we found empirically that Gurobi took over five minutes} to produce a tiling solution for one character shown in Figure~\ref{fig:teaser}.
Our goal is to resort to a data-driven approach\phil{,} which can produce very good, while not necessarily optimal, solutions at much greater efficiency. 

%\phil{I avoid saying our tiling problem here, since we avoid continuous tile placements}
In general, the solution space for tiling is immense, spanning both a discrete search for tile type selection
and arrangement, as well as a continuous search for tile orientation and positioning. 
Also,
%In addition, 
we must respect hard
constraints related to input boundary, holes, and tile overlaps. A traditional approach to \phil{solving} the problem would search 
over the space of tile selection and placement, progressively laying out tile instances to fill the tiling region, possibly 
with trial-and-error and backtracking in the solution search. However, the ensuing computation would be prohibitively 
expensive.

By taking a learning-based approach, we train a deep neural network, whose parameters 
%can 
encode knowledge or
{\em patterns\/} of the tiling solutions. At test time, the learned patterns would guide the tile placement, instead of
relying on expensive on-line search. Specifically, we model our tiling problem as an instance of {\em graph learning\/}.
In our graphs, nodes represent candidate tile locations in the solution space and edges encode tile overlap relations 
and regular tile connections. Such a representation allows us to 
%make use of 
adopt a {\em graph neural network\/}, which is
coined TilinGNN, to process features on graph nodes. 
%Specifically, 
These features are aggregated along graph edges,
via {\em graph convolution\/}, to predict 
%candidate 
tile locations for inclusion in the 
%tiling 
solution.

Structure-wise, TilinGNN has a two-branch network architecture with a series of neighbor aggregation 
modules and overlap aggregation modules to learn features related to tile connection and overlap. In such a way, 
we train TilinGNN to learn to maximize the tiling solution coverage, while avoiding tile overlaps and holes.
Importantly, we formulate these tiling criteria as loss terms, which are functions of the network outputs. Hence, 
our network can be trained in a {\em self-supervised\/} manner, without the need for any human-provided tiling 
solutions as ground truth during \phil{the} training.
%In the end, we can make use of a trained TilinGNN to help arrange tiles and produce tiling solutions.}

%Our GNN is trained to {\em predict
%selection probabilities\/} for the tiles, leading to tile selection and placements.
%Also, we formulate three self-supervised loss terms to guide TilinGNN to learn to maximize the tiling coverage and to avoid overlaps and holes in the tiling solutions.}
%
%\phil{I can't follow the meaning of this sentence:}
%While the graph structures 
%which correspond to tiling solutions for different shapes do not appear to exhibit any global consistency, we have found 
%that these solutions do possess learnable {\em local\/} %\xh{seamless} 
%patterns which can be exploited by a {\em graph neural network\/}, or GNN.
%Importantly, the network can be 
%{\em self-supervised\/}, without needing any human-provided tiling solutions. Our GNN is trained to {\em predict
%selection probabilities\/} for the tiles, leading to tile selection and placements.
%
%
%\todo{RZ: Maybe provide more details.}
%\phil{done!}

The runtime of using a trained TilinGNN model to tile a shape is roughly linear to the number of candidate tile locations.
Typically, we can generate a tiling solution in less than a minute, which is \final{hard to achieve} using tiling approaches 
based on traditional combinatorial search. To verify this, we perform experiments to compare our method with COIN-OR/BC~\shortcite{coin-bc} and \final{Gurobi~\shortcite{gurobi}} for tiling different shapes of varying sizes.
The results confirm the superior speed and tiling quality of our method.
Besides, we showcase a variety of tilings produced by our method using assorted tile sets with single or multiple tile types (see Figure~\ref{fig:teaser}), present an interactive design tool, compare our results with manual solutions, and present various evaluations on the network architecture and loss.
%to show the quality of our method.}
%\phil{Let's come back to here after Sec 6 is drafted...}
%
%\todo{RZ: One paragraph to summarize results, e.g., instantaneous, generality, multiple tiles, refer to Figure~\ref{fig:teaser}.}

To the best of our knowledge, TilinGNN represents the first attempt at generating non-periodic tilings using a deep neural network. Since this form of tiling is only one particular instance of the more general problem setting of ``learning to select'' over a graph structure, we believe that our approach can bring insights and open up new directions for solving other similar combinatorial problems in computer graphics research, e.g., the design of hybrid meshes
%with triangles and quads
~\cite{Peng-2018-Hybrid-Meshes} and various computational assembly problems~\cite{luo-2012-Chopper,masonry-2014-Assembly,xu-2019-technic-lego}.
%\todo{add one more ref}
%\phil{Anything else?}.
%Scott-2005-Triangle-Quad,

%Richard: Btw, Philip, I would have done a completely different intro. The paper won't be attractive if it is just a tiling paper. I was thinking to start with how ML has impacted graphics, and the typical use of ML for typical problems, and then ML to solve hard optimizations, and only then tiling. What do you think?
%%
%To sum up, these are very challenging problems, with not only an immense search space but also a nonlocal characteristic.
%Having said that, a local decision in the result could affect the global structure, and hence, the overall results.
%
%In this work, we revisit the tiling problem and explore a novel machine learning solution by learning to tile through self supervision.
%\phil{a statement to define our overall goal... Then our target}
%%
%The key insight in our approach is that when the machine sees a domain to be tiled, we encourage it to learn to ...

\section{Related work}
\label{sec:related}
%In the past decade, we witness how machine learning impacts computer graphics research and enables the community to achieve improved and new results in many existing difficult problems.
%

Generally speaking, machine learning enables a computer system to perform tasks without pre-defining model features 
or providing explicit instructions. Instead, the features and computational parameters behind the actions are 
{\em automatically learned\/} from data rather than being fixed or handcrafted.
Most deployments of machine learning techniques to computer graphics can be found in image and video 
processing, e.g.,~\cite{bau-2019-photo-manipulation,sun-2019-portrait-relighting,jamriska-2019-stylizing-video}, where 
% P: , here to follow the US style
conventional convolutional processing over regular grid data are typically applied. Recently, some success has been 
achieved in geometric deep learning over irregular data such as shapes~\cite{li-2017-grass, mo-2019-structurenet} 
and scene structures~\cite{wang-2019-PlanIT, li-2019-grains, gao-2019-deep-meshgen}, which are represented as attributed
trees or graphs. What is common about these methods is that they are {\em supervised\/}, where the networks were trained
to learn structural {\em consistencies\/} in a shape/scene collection arising from shared semantic or functional attributes.

In contrast, our goal is to develop a {\em self-supervised\/} tiling solution that works on arbitrary input shapes 
without assuming any shared commonality among them. In this section, we discuss relevant works 
to the tiling problem, including image and pattern generation, shape decomposition, and 3D shape assembly. A coverage 
on neural combinatorial optimization models is also provided.

%%%%%%%%%%%%%%%%%%%%%%%%%%%%%%%%%%%%%%%%%%%%%%%%%%%%
\ifdefined\negativevspace
\vspace*{-3pt}
\fi
%\paragraph{Computational tiling.}
\paragraph{Tiling for image and pattern generation.}
Artists, graphics designers, and mathematicians have been interested in the problem of tilings and their properties for centuries.
Theories and in-depth analysis of most instances of the tiling problem can be found in~\cite{grunbaum-1986-tilings-book}.
Existing works focus mostly on filling the 2D plane using a small set of {\em fixed\/} shapes as the tiles, for example, the periodic and 
aperiodic tilings, and polyomino tilings.
%, Penrose tilings, 
%
In computer graphic research, various 
%computational design 
tools have been developed to help produce different kinds of intriguing 
tiling patterns. For example, the problem of Escherization~\cite{kaplan-2000-escherization},
creation of decorative mosaics using colored tiles~\cite{hausner-2001-decorative-mosaics,smith2005animosaics},
texture generation with Wang tiles~\cite{stam-1997-aperiod,cohen-2003-WangTile},
\final{object distribution on the plane~\cite{hiller-2003-obj-plane}},
the construction of traditional Islamic star patterns~\cite{kaplan-2004-star-patterns},
\final{and the design of artistic packing layouts~\cite{reinert-2013-artistic-packing}.}
%Animated mosaics~\cite{smith2005animosaics}.
%
These works focus either on the mathematics to support the tiling constructions, or on methods with heuristics such as 
Voronoi Diagrams to divide the plane and fill it with tiles.

\ifdefined\negativevspace
\vspace*{-3pt}
\fi
\paragraph{Decomposition.}
A line of works that bears some resemblance to tiling follows the principle of {\em design by composition\/}. However, the ensuing 
{\em decomposition\/} problem is characteristically different from tiling, since the compositional elements are not of fixed shapes; they
typically share some common geometric or appearance properties, but are otherwise of different shapes, even deformable.  Also, there is
often a desire to minimize the decomposition size.
%size of the decomposition. 
Literatures on decomposition are vast.
Some 
%of the
%, instead of being replicates of a tile set.
exemplary works include
%jigsaw 
image mosaics~\cite{kim-2002-JIM,xu-2019-DISC},
shape collages~\cite{Huang-2011-Arcimboldo-like,Kwan-2016-2DCollage}, convex decomposition~\cite{lien-2004-ACD},
%$k$-set tilable surfaces~\cite{fu-2010-Kset}, This is 3D but others are 2D
ornamental packing~\cite{Saputra-2017-OrnamentalPacking},
fabricable tile decors~\cite{chen-2017-tile-decors},
approximate dissection~\cite{Duncan-2017-ApproximateDissection}, and 
reversible linked dissection~\cite{Li-2018-ReversibleDissection}.
%\phil{are you sure that all of the tiles in these works are deformable? I haven't checked every single one}

%%%%%%%%%%%%%%%%%%%%%%%%%%%%%%%%%%%%%%%%%%%%%%%%%%%%
\ifdefined\negativevspace
\vspace*{-3pt}
\fi
%\paragraph{3D shape assembly.}
\paragraph{\final{Assemblies of fixed blocks.}}
Assembling a shape using {\em a small set of building blocks\/} 
%Ph: perhaps highlight these words; otherwise, there will be a huge body of works from many
may also be 
%considered 
viewed as a form of tiling.
%In computer graphics, many works on computational assembly problems are also in the category of tiling.
%
Eigensatz et al.~\shortcite{eigensatz-2010-paneling}, Fu et al.~\shortcite{fu-2010-Kset}, and Singh and 
Schaefer~\shortcite{singh-2010-triangle} independently develop methods for the problem of constructing 
a small set of shapes (or panels) that together can tile a given 3D surface.
%
%\cite{fu-2010-Kset} generated a set of quads which instances can produce a tiled surface, approximating the input surface.
\final{Peng et al~\shortcite{peng-2014-layout} tackle the problem of tiling a domain with a set of deformable templates.}
Luo et al.~\shortcite{luo-2015-legolization} account for the shape, colors, and physical stability when their method searches for LEGO brick constructions.
Skouras et al.~\shortcite{skouras-2015-interlocking-elements} develop a design tool to build structures from interlocking quadrilateral elements.
Chen et al.~\shortcite{chen-2018-PreFab} compute an internal core built by universal blocks to support a 3D-printed shell, 
Xu et al.~\shortcite{xu-2019-technic-lego} search for 3D LEGO Technic brick constructions, given the user-input sketches,
\final{while Peng~\shortcite{peng-2019-checkerboard} design 2D checkerboard patterns with black rectangles derived from quad meshes.}
%\final{Compared with our formulation, tiles in Peng~\shortcite{peng-2019-checkerboard} are connected with full edges, while our formulation considers also partial edge connections and it needs to explicitly avoid holes in the computation.}
%
\final{In general, computational assembly problems, including ours, are combinatorial by nature, so they are typically solved by integer programming (IP) or heuristic search.
Being general tools for combinatorial problems, IP solvers could be rather time-consuming for large problems.
On the other hand, specifically-designed heuristic search could perform fast, but it may not be general for solving problems of different forms.
%While it is possible to adopt these approaches for our tiling problem, we need to overcome their shortcomings.
}
%
%\phil{the logic is not clear in the last two sentences... the word "on the other hand" starts a bit awkward}
%
%Search-based methods, at one extreme, can perform fast using specifically-designed heuristic; and at the opposite extreme, will be very time-consuming, given an immense search space.
%by exhaustive search.
In our work, we \final{trade-off the advantages of the two approaches, and leverage} a graph neural network to learn the solution patterns \final{as heuristics,}
 %of a specific kind of combinatorial optimization problem (i.e., tiling) 
\final{so that} we can make \final{fast} predictions using the trained model.

\ifdefined\negativevspace
\vspace*{-3pt}
\fi
\paragraph{Graph neural networks (GNNs).}
A natural extension from regular grid processing to 
%dealing 
deal %\phil: just to tighten the words
with irregularly structured data is to utilize graph representations. 
There has been much recent development on deep learning using graph neural networks (GNNs)~\cite{wu-2019-GNNSurvey},
in particular, graph convolutional networks (GCNs), which our method adopts. GCNs build stacked convolutional layers
over graph structures to 
%which 
enable feature aggregation based on neighborhood information.
%in the graphs.
%
Some recent works from computer graphics apply GCNs in different contexts. In MeshCNN, Hanocka et al.~\shortcite{hanocka-2019-meshcnn} 
perform graph convolution and pooling over mesh edges. In PlanIT, Wang et al.~\shortcite{wang-2019-PlanIT} develop
a GCN-based generative model for indoor scenes, where the network operates on scene graphs that
encode object-to-object relations. 
In StructureNet, Mo et al.~\shortcite{mo-2019-structurenet} encode 3D shapes using $n$-ary graphs and consider both part geometry and inter-part relations in the network training.
In our work, GCNs are applied in a novel way for tiling, where the graph encodes the tile overlap and connection relations, 
and the network is trained to predict probabilities for tile selection and placement.
%\phil{no s after placement?}

%%%%%%%%%%%%%%%%%%%%%%%%%%%%%%%%%%%%%%%%%%%%%%%%%%%%
\ifdefined\negativevspace
\vspace*{-3pt}
\fi
\paragraph{Neural combinatorial optimization.}
Recent works on neurally guided optimization explore data-driven and learning-based schemes for solving discrete and 
combinatorial problems. 
% Phil: renowned -> notable, just for avoiding a single word at the end of the paragraph
A typical and notable example is Alpha Go~\cite{silver2017mastering}, which trains a machine
to predict the best next move in Go play using Monte Carlo tree search.
%, which learned to the disc placements on a fixed-sized game board. % in an unsupervised manner.
Many works attempt to use machine learning to efficiently solve classical NP-hard problems such as 
traveling salesman~\cite{vinyals-2015-pointer-networks, bello-2016-neural-combinatorial, dai-2017-combinatorial-graphs},
boolean satisfiability~\cite{yolcu-2019-local-search},
maximum independent set~\cite{li-2018-treesearch,abe-2019-nphard-reinforcement}\final{, graph coloring~\cite{lemos-2019-graph-color}}, and 
maximum cut~\cite{dai-2017-combinatorial-graphs,yolcu-2019-local-search}. The particular learning-based mechanisms 
include recurrent neural networks, attention models, deep reinforcement learning, as well as GCNs.

The common insight of these works is to train a \final{neural} network to predict selection probabilities over discrete options, and use a simple 
selection strategy (e.g., greedy) to incrementally construct a solution, as guided by the probabilities. Our work inherited the same 
spirit, where we model the tile placements as discrete options, but differs in the way of the learning process. To train a network, 
we neither have a data set of tiling solutions to supervise the training, nor train in a reinforcement learning framework via trials 
and errors. Instead, we formulate losses to evaluate the quality of the network output, training the network in a self-supervised manner.

\section{Overview}
\label{sec:overview}
%%%%%%%%%%%%%%%%%%%%%%%%%%%%%%%%%%%%%%%%%%%%%%%%%%%%
%\input{figures/overview}
%%%%%%%%%%%%%%%%%%%%%%%%%%%%%%%%%%%%%%%%%%%%%%%%%%%%%%%%%%%%%%

\begin{figure}[t]
	\centering
	\includegraphics[width=0.9\linewidth]{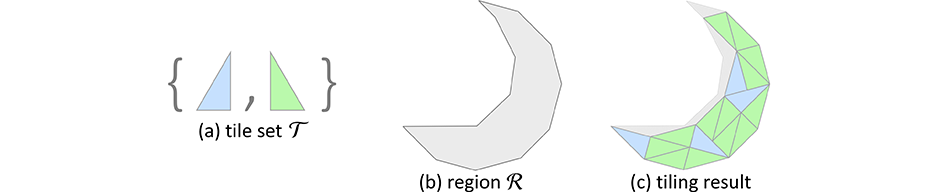}
	\vspace*{-1mm}
	%\ifdefined\negativevspace
	%\vspace*{-1.5mm}
	%\fi
	\caption{The tiling problem.
		Given a tile set $\mathcal{T}$ (a) and a 2D region $\mathcal{R}$ to be filled (b), we aim to produce a tiling (c) that maximally covers the {\em interior\/} of the given region without overlap or hole between the tile instances.}
	\label{fig:problem}
	%\ifdefined\negativevspace
	%\vspace*{-0.5mm}
	%\fi
	\vspace*{-1mm}
\end{figure}

%\vspace*{-3pt}
\paragraph{Problem definition.}
Given a set $\mathcal{T}$~of a few simple 2D shapes representing the input tile types, and a connected region $\mathcal{R}$ in 2D, we aim to arrange instances of tiles from $\mathcal{T}$~over $\mathcal{R}$ without overlap or hole between the tile instances, such that the resulting tiling can maximally cover the interior of $\mathcal{R}$ without exceeding its boundary. Figure~\ref{fig:problem} shows an illustration of the problem.

Note that we can rotate and translate the tiles, but not flip or scale them, when arranging the tile instances.
%\xh{In most cases, both $T_i$ and $\mathcal{R}$ can be represented by simple polygons, but $\mathcal{R}$ can also contain hole(s) inside, see the letter ``R'' in Figure~\ref{fig:teaser} for an example.}
%On the other hand, both $T_i$ and $\mathcal{R}$ are 2D polygons that can be represented by an ordered list of 2D coordinates in counter-clockwise order.
Since \final{the shape of $\mathcal{R}$ is arbitrary} and the tile set is limited, \final{completely filling the entire} $\mathcal{R}$ may not be always possible, as demonstrated by the example shown in Figure~\ref{fig:problem}(c). Last but not least,
the region $\mathcal{R}$ itself may contain holes, as shown by \final{characters ``A'' and} ``R'' in Figure~\ref{fig:teaser}.

Overall, the problem setting is quite general, covering an infinite number of problem instances.
When considering different tile sets and tiling regions as inputs, the difficulty of the problem varies.
For example, tiling a general 2D region using trominoes has been proved to be NP-complete~\cite{moore-2001}, 
while tiling a square region using smaller squares is relatively easy.
\ifdefined\negativevspace
\vspace*{-3pt}
\fi
\paragraph{Challenges.}
%There are three major challenges to our tiling problem.
%
First, the search space is immense, as we must handle three closely-coupled sub-problems:
(i) which tile to pick;
(ii) how to position and orientate each tile; and 
%the position and orientation of each tile; and 
(iii) how many instances of each tile type to use in the tiling.
The decision variables of problems (i) \& (iii) are discrete, while those of problem (ii) are continuous.
%Also, the constraints on these variables are non-linear, since we need to evaluate the overlaps and adjacency between the tile instances in the results.
%with irregular shapes.
%requiring an efficient evaluation of tile overlaps and adjacency between tiles.

Second, the avoidances of overlaps, holes, and tiles 
%touching 
surpassing
the input shape boundary are all hard constraints. In 
general, we need to actively evaluate the tile placements against these constraints and discard those that violate the
constraints; this can be exceedingly time-consuming given the immense search space.

Finally, tile placements require {\em both local and global considerations\/}.
Locally, the placement of a tile is closely related to the placements of the neighboring tiles, so that we can avoid the tile overlap and minimize the gaps between tiles to maximally cover $\mathcal{R}$.
Globally, a change \final{in} a local tile placement can subsequently affect its neighbors and even the entire tiling, since the hard constraints can propagate over the neighbors successively over the tiling.

%%%%%%%%%%%%%%%%%%%%%%%%%%%%%%%%%%%%%%%%%%%%%%%%%%%%

%%are non-linear\xh{, due to the irregular shape and different orientation of the polygons.
%Hence, directly formulating a problem to optimize the pose of every tile under such constraint is improper.
%To meet them, we 
%%have not only complicate the computation?
%%Also, the constraints on these variables are non-linear, since we 
%have to appropriate model the problem, such that the evaluation of the constraints on the variables are straightforward, easy to be qualified, and %suitable for optimization.}
%%have to actively evaluate the overlaps and holes between the tile instances when our method generates the tiling results.
%
%%To meet these constraints require the method not only to handle 
%\phil{To XH: anything else that we may elaborate, particularly on the last item}

%Third, the constraints on a tiling solution are hard and strict. 
%Even for a just qualified solution, we need to consider not only avoiding the overlap for solution feasibility, but also avoiding holes inside a tiling %for usability or aesthetics.
%Violating a little bit will result in an unacceptable solution, not to mention solution optimality.

%\\part{phil}{I stop here... to work on the next paragraph, I need to see a better (and completed) Figure 3 and figure caption to describe what's going on inside the figure}

%%%%%%%%%%%%%%%%%%%%%%%%%%%%%%%%%%%%%%%%%%%%%%%%%%%%%%%%%%%%%%
\ifdefined\negativevspace
\vspace*{-3pt}
\fi
\paragraph{Our approach.}
In this work, we explore a new approach by designing and training a graph neural network with self-supervised 
losses to {\em learn\/} to predict tile placement locations. Then, at test time, we can make use of the 
network to generate tiling efficiently. Altogether, our approach has three phases, as illustrated in Figure~\ref{fig:overview}:
\begin{itemize}
\item[(i)]
Given \final{an} input tile set $\mathcal{T}$, we first pre-compute a superset of tile configurations \final{that} enumerate\final{s} all candidate tile placement locations for $\mathcal{T}$; see Figure~\ref{fig:overview}(a) and Section~\ref{sec:modeling}.
\item[(ii)]
Our next goal is to train a GNN, coined TilinGNN, to learn to predict tile placement locations for the given tile set; see Figure~\ref{fig:overview}(b). To do so, we generate a random 2D shape to crop a tiling configuration so as to locate 
candidate tile locations inside the shape. Then, we create a graph structure to describe the adjacency between 
candidate tile locations in the shape and train TilinGNN to predict tile placement locations by formulating self-supervised 
losses to avoid overlaps and holes in the tiling. We repeat this process using many different random shapes, as a 
means for data augmentation, to facilitate the training of TilinGNN; see Section~\ref{sec:method}.
\item[(iii)]
To fill a test region, we first locate candidate tile locations within the region, then fill the region by using the TilinGNN trained on the target tile set to progressively predict tile placements; see Figure~\ref{fig:overview}(c) and Section~\ref{ssec:test} for details.
\end{itemize}
%
%Once our TilinGNN is trained, using it to tile a region \rz{at test time} is almost {\em instantaneous\/}.
%Hence, we further design an interactive interface to enable us to modify the boundary of the tiling region and explore different tiling results %accordingly; see Section~\ref{sec:results}.
%\TODO{Revisit}
%%\phil{NOTE: revisit or remove this paragraph, after drafting sec 6... it may not be good to emphasize this}

%\phil{I stop here... please fill in the blank and fix all my comments for you... for Fig.4, shall we remove the seed tile? This is an implementation thing I guess so we can talk about this only in Section 4, right?} \xh{Yes, I revised.}

%To cope with the large and irregular search space, we first locate a finite number of tile placements inside an input tiling region (Figure~\ref{fig:overview} (i) \& (d)).
%Then, we model the tiling generation problem as selecting a subset of the candidate tiles to fulfill the tiling requirements.
%We encode the tile placements and their spatial relation into an {\em adjacency graph} (Figure~\ref{fig:overview} (e)), which is sent to a graph neural network to make the node selection on the adjacency graph and output the probability of selecting each node.
%where each node represents a tile placement and edges connecting these nodes record the adjacency relation between tiles.
%The trained model will then be used to guide a simple probability-based selection procedure to select tiles, where the network may be called several times with the unselected candidate tiles as input, see Figure~\ref{fig:overview} (j), (k) \& (l).

\section{Modeling the Tiling Problem}
\label{sec:modeling}

To adopt a neural network to perform tiling is non-trivial.
First, the input data, i.e., both the tiling region $\mathcal{R}$ and the tiles in $\mathcal{T}$, have irregular shapes, so we cannot directly process them by conventional convolutional neural networks.
Second, the tiling problem can have many different problem instances, when using different tile sets to tile different kinds of regions.
The neural network architecture should be general to handle them in a unified fashion.
Lastly, the number of tiles in the input tile set is not fixed, whereas the number of tiles in the output tiling is unknown.
The approach should be able to allow such flexibility for both the inputs and outputs.

Hence, we approach the problem by first modeling the tile set and enumerating the candidate tile locations (Section~\ref{ssec:tile_set}).
Then, we model the problem of tiling generation as a graph problem and present necessary constraints and objectives (Section~\ref{ssec:enc_graph}).
These are preparation works to later enable processing by TilinGNN.

%In those computational tasks on images, video, game board, or point cloud, the data keeps the same structure in the computation process as it is originally organized.
%The data in our computational tiling problem, however, requires non-trivial encoding to be processed. 
% due to the universal problem setting, the number of polygons in the inputs and output may also vary.
%Although simply using a list of vertex positions is enough to reproduce all information of a polygon, but it lacks a direct representation of the spatial occupation inside the polygon.

%%%%%%%%%%%%%%%%%%%%%%%%%%%%%%%%%%%%%%%%%%%%%%%%%%%%

\subsection{Modeling the Tile Set and Tile Placements}
\label{ssec:tile_set}
\begin{figure}[t]
	\centering
	\includegraphics[width=0.9\linewidth]{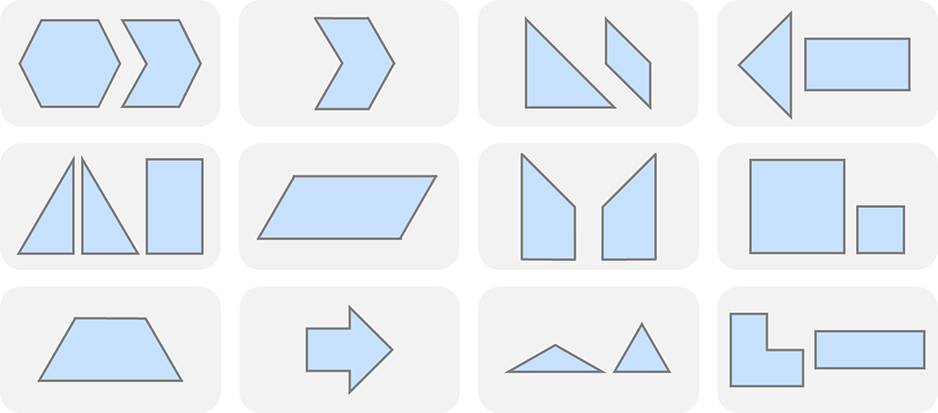}
	\ifdefined\negativevspace
	\vspace*{-1.5mm}
	\fi
	\caption{Exemplary tile sets.
	Some of them have specific names, e.g., the rightmost two at the bottom are the labyrinth tiles and tromino tiles.}
	%\caption{Exemplary tile sets.\TODO{revisit: (bottom rightmost)!!!}}
	\label{fig:tile_set}
	\ifdefined\negativevspace
	\vspace*{-1.5mm}
	\fi
\end{figure}

First, we state two fundamental requirements on the tile set $\mathcal{T}$:
\begin{itemize}
\item[(i)]
$\mathcal{T}$ should {\em seamlessly tile the plane\/}, such that the tile instances \final{may} fill target region $\mathcal{R}$ without holes and tile overlaps; and
%, in order to guarantee the existence of the tiling solution.
%Also, to better tile the input region, which may have different shapes, the possible arrangements of tiles in a tile set are better to be more than periodic.
%Although the problem of telling whether a finite set of tiles can tile the infinite plane is undecidable in general~\cite{robinson-1971-undecidability}, there are still a large number of shapes or shape collections was found to be managed to tile the 2D plane with specific arrangements.
%Examples of these tiles can be found in ~\cite{grunbaum-1986-tilings-book}.
%
\item[(ii)]
the tile set and tile connection rules \finaltwo{together should induce a {\em finite number\/} of candidate tile locations inside a finite region. Indeed, since TilinGNN is a {\em selection\/} network, it has to operate on a finite number of candidate choices. % (i.e., candidate tile locations).
The candidate tile locations, or the superset, should then form a {\em periodic grid\/}, e.g., see Figure~\ref{fig:overview}(a). However,
%Note that even though the grid is periodic, 
the generated tilings, formed by selected candidate tiles, can be far from periodic.}
%\phil{Note that I tried to write a bit more to make it easier to follow and read}
%
%
%\xh{infinite}
% amount of {\em candidate tile locations\/} \xh{in a finite region}.
%\xh{Because our network learns to select from the set of {\em all} candidate tile placements, thus the set should neither be infinite nor incomplete.}
%
% It is because TilinGNN is a graph neural network, its computation is limited by the memory available on the GPU.
%Yet, TilinGNN can already handle a variety of tile sets, as exemplified in Figure~\ref{fig:tile_set}.
%
%Draft from XH:
%Second, although every tile can be placed on arbitrary locations in general, the hard constraints on overlaps and holes make tiles can only be placed in a restricted amount of possible locations.
%The possible locations of tiles induced by a tile set should be manageable and computable in a modern computer.
%
\end{itemize}
Figure~\ref{fig:tile_set} shows some of the tile sets supported by TilinGNN.

%%%%%%%%%%%%%%%%%%%%%%%%%%%%%%%%%%%%%%%%%%%%%%%%%%%%
%%%%%%%%%%%%%%%%%%%%%%%%%%%%%%%%%%%%%%%%%%%%%%%%%%%%
%%%%%%%%%%%%%%%%%%%%%%%%%%%%%%%%%%%%%%%%%%%%%%%%%%%%
\begin{figure}[t]
	\centering
	\includegraphics[width=0.95\linewidth]{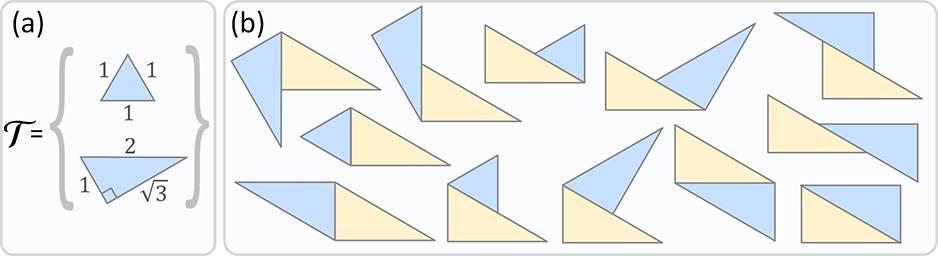}
	\vspace*{-1.5mm}
	%\ifdefined\negativevspace
	%\vspace*{-1.5mm}
	%\fi
	\caption{For the example tile set given in (a), there are twelve possible ways to place \final{a neighbor} (b) around the right-angled tile marked in yellow.}
	%	\caption{We show all possible tile neighbors (b) around the yellow tile instance with edge connection and corner-aligned. 
	%		The neighbors are instances of a given tile set (a).}
	\label{fig:adj_tiles}
\end{figure}
%%%%%%%%%%%%%%%%%%%%%%%%%%%%%%%%%%%%%%%%%%%%%%%%%%%%

%\phil{comment out, if this is exactly at the page break}
%\vspace*{-3pt}
\paragraph{Tile neighbors.}
In general, the neighbor of a tile can be placed at any location next to the tile, as long as they contact by an edge or a point.
In this work, we consider only neighbors that contact by a {\em common edge segment\/} and do not allow the neighbor to continuously slide over the contacting edge, since this \phil{induces} an infinite number of candidate tile placements (see top-left case in inset figure below).
Besides sharing a full edge (see top-middle case in inset figure), if a 
\setlength{\columnsep}{2.8mm} % wrapfigure margin
\begin{wrapfigure}{r}{0.43\columnwidth}
\vspace{-3pt}
%\centering
\includegraphics[width=0.42\columnwidth]{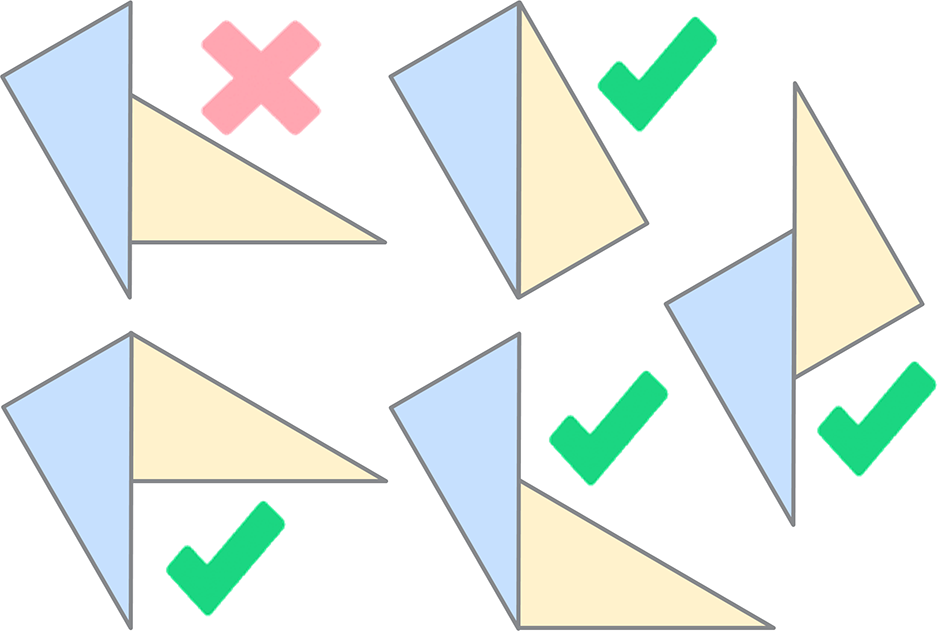}
\hspace{-30pt}
\vspace{-12pt}
\end{wrapfigure}
side length of a tile is a multiple of some shorter side lengths in the whole tile set, the tiles can then connect with ``quantized'' lengths along the shared edge (see bottom three cases in inset figure).
See also Figure~\ref{fig:adj_tiles}, for more examples of the neighbor tile placement locations.
%\xh{We also allow the two triangles to connect by shares half of their longest edge, because the length of the shared segment is just equal to the shortest edge. }

%XH's draft:
%Tiles in a seamless tiling solution connect with neighbor tiles by either sharing {\em edges}, or a {\em corner} on their respective boundary, we name %them edge and corner connection, respectively.
%We can only consider the edge connection between neighbor tiles without ignoring the second situation, because a corner-connected neighbor of a tile can %be always modeled as an edge-connected neighbor of another tile.
%Note that there is an infinite number of ways to connect two tiles by sharing edge, because the two tiles can continuously shift along a straight shared %edge.
%We discretize the continuous connection by considering only the situation where two tiles are {\em corner-aligned}; see the example in the inset figure. 
%Moreover, we define two additional rules to reduce the exploration space in our tiling problem:
%First, the length of the shared edge of one of the two neighbor tiles should be the integer multiple of the other; otherwise, the unshared portion of the %edge can probably not connect with other tile instances with an edge of the same length.
%Second, there should be no overlap or holes between a tile and its neighbor. This situation may happen for concave tiles.
%Figure~\ref{fig:adj_tiles} shows all the neighbors of the yellow triangle under the rules.

\begin{figure}[t]
	\centering
	\includegraphics[width=0.95\linewidth]{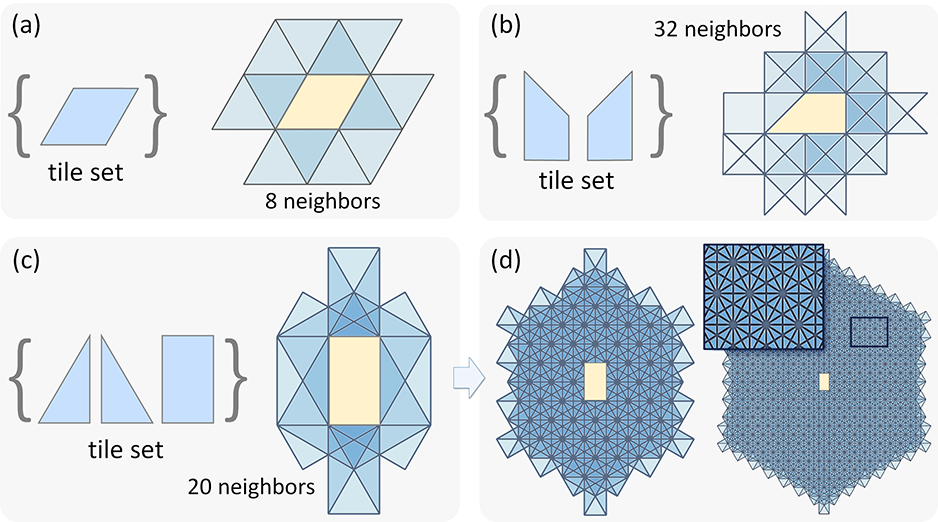}
	\vspace*{-1.5mm}
	%\ifdefined\negativevspace
	%\vspace*{-1.5mm}
	%\fi
	\caption{(a)-(c) Given a tile set, we pick one of its tiles as \final{the} seed (yellow) and locate all surrounding neighbors, which are rendered in transparent altogether.
		(d) If we repeat the process recursively with the neighbors, we can find the candidate tile placement locations over larger regions, e.g., 
		there are 742 (left) and 5301 (right) tile locations in these two results.}
	%
	%	\caption{Given a tile set (a), we show 1, 3, 7 number of rings neighbors from a seed yellow tile. The number of tiles in these figures are: 21, 742, 5301, respectively.}
	\label{fig:super_graphs}
	%\vspace*{-1mm}
\end{figure}

\ifdefined\negativevspace
\vspace*{-3pt}
\fi
\paragraph{Candidate tile placements.}
Based on the above tile connection rule,
%and also the second fundamental requirement, 
we can determine {\em a finite set of candidate tile placement locations\/} over a given region.
We denote such a set as $\mathcal{P}$.
To find $\mathcal{P}$ for a given tile set, one way is to pick one of the tile(s) in the tile set as a seed, locate all possible neighbors around the seed (see Figures~\ref{fig:super_graphs} (a)-(c) for examples), then repeat the process recursively with the neighbors, until we fill the target region (see Figure~\ref{fig:super_graphs}(d)).
Besides, for the \final{tromino} tile set shown at the bottom right corner of Figure~\ref{fig:tile_set}, we can simply sweep each tile type over a \final{regular} 2D grid for each unique orientation of the tile type to find $\mathcal{P}$ \final{of} the tile set.

Clearly, the size of $\mathcal{P}$ grows with the region size and the complexity of the tile set, we thus limit the size of $\mathcal{P}$ by the available GPU memory.
Typically, our current implementation of TilinGNN on a single GPU can work with $\sim$5k (candidate) tile locations.

%%%%%%%%%%%%%%%%%%%%%%%%%%%%%%%%%%%%%%%%%%%%%%%%%%%%
\begin{figure}[t]
	\centering
	\includegraphics[width=0.99\linewidth]{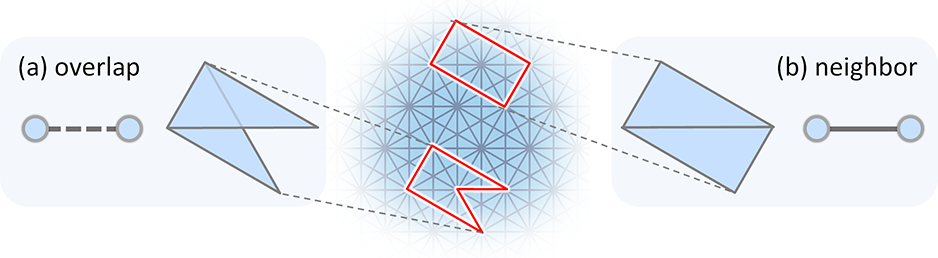}
	\ifdefined\negativevspace
	\vspace*{-3mm}
	\fi
	\caption{The two types of edges in the adjacency graph:
		(a) an overlap edge (dashed line) between nodes, of which the tile locations overlap; and
		(b) a neighbor edge (solid line) between nodes, of which the tile locations contact.}
	\label{fig:enc_graph}
	%\vspace*{-1mm}
\end{figure}
%%%%%%%%%%%%%%%%%%%%%%%%%%%%%%%%%%%%%%%%%%%%%%%%%%%%

%%%%%%%%%%%%%%%%%%%%%%%%%%%%%%%%%%%%%%%%%%%%%%%%%%%%

\subsection{Modeling Tiling as a Graph Problem}
\label{ssec:enc_graph}
By means of modeling the candidate tile placement locations, {\em a tiling problem can be cast as a graph problem\/} for TilinGNN to work on.
That is, given a region $\mathcal{R}$ to tile, we can first locate a set of candidate tile placement locations in the region.
We can then create an adjacency graph (denoted by $\mathcal{G}$) to describe the connectivity between candidate tile locations by regarding each candidate tile location as a graph node, and construct edges for two cases:
\begin{itemize}
\item[(i)]
If two candidate tile locations overlap each other, we construct an {\em overlap edge\/} to connect their respective nodes in $\mathcal{G}$; and
\item[(ii)]
If two candidate tile locations contact each other by an edge segment without any overlap, we construct a {\em neighbor edge\/} to connect their respective nodes in $\mathcal{G}$.
\end{itemize}
Figures~\ref{fig:enc_graph} (a) \& (b) illustrate these two types of edges.
We use dashed lines and solid lines to denote overlap and neighbor edges, respectively; see also the adjacency graph example in Figure~\ref{fig:overview}(b).

Therefore, adjacency graph $\mathcal{G}$ is an undirected graph that can be written as $\mathcal{G} = \{ \mathcal{V} , \mathcal{E}_\text{ovl} , \mathcal{E}_\text{nbr} \}$ with
node set $\mathcal{V}$,
overlap edge set $\mathcal{E}_\text{ovl}$, and
neighbor edge set $\mathcal{E}_\text{nbr}$.
In this way, we can re-formulate our tiling problem as the problem of
\vspace{1mm}
\begin{quote}
{\em Finding a (maximum) subset of nodes in $\mathcal{G}$, such that all adjacent nodes are connected by edges in $\mathcal{E}_\text{nbr}$ and no two nodes are connected by any edge in $\mathcal{E}_\text{ovl}$.}
\end{quote}
\vspace{1mm}

%%%%%%%%%%%%%%%%%%%%%%%%%%%%%%%%%%%%%%%%%%%%%%%%%%%%
\begin{figure*}[t]
	\centering
	\includegraphics[width=0.99\linewidth]{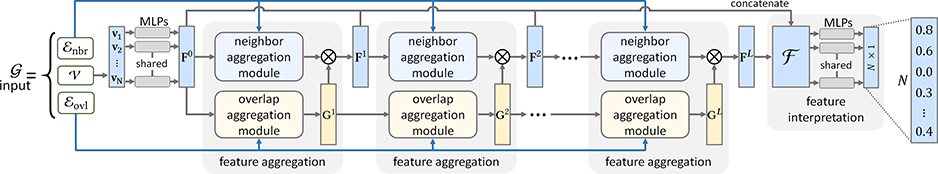}
	\ifdefined\negativevspace
	\vspace*{-1.5mm}
	\fi
	\caption{The overall architecture of TilinGNN is a two-branch graph convolutional neural network. The network progressively propagates and aggregates node features ($\mathbf{F}^l$ and $\mathbf{G}^l$) over the neighbor and overlap edges in an adjacency graph, using the neighbor aggregation and overlap aggregation modules (see Figure~\ref{fig:graph_conv}) to predict node selection probabilities for candidate tile placements in tiling generation.
	$N$ is the total number of nodes in the graph;
	both $\mathbf{F}^l$ and $\mathbf{G}^l$ are node features of dimensions $N \times C$;
	$C$ is the number of feature channels; 
	$L$ is the number of layers in TilinGNN; and
	{\Large $\otimes$} is the element-wise product.}
	\label{fig:net_architec}
	\ifdefined\negativevspace
	\vspace*{-1mm}
	\fi
\end{figure*}
%%%%%%%%%%%%%%%%%%%%%%%%%%%%%%%%%%%%%%%%%%%%%%%%%%%%

\final{Here, }we aim to find a tiling that maximally \final{covers} tiling region $\mathcal{R}$ without holes and tile overlaps.
Such a problem can be further cast as an optimization, in which we select a subset of nodes in $\mathcal{G}$, such that
(i) the total area of the tiles associated with the selected nodes is maximized;
(ii) we aim to avoid holes between tiles, by maximizing 
\setlength{\columnsep}{2.8mm} % wrapfigure margin
\begin{wrapfigure}{r}{0.38\columnwidth}
\vspace{-10pt}
%\centering
\includegraphics[width=0.37\columnwidth]{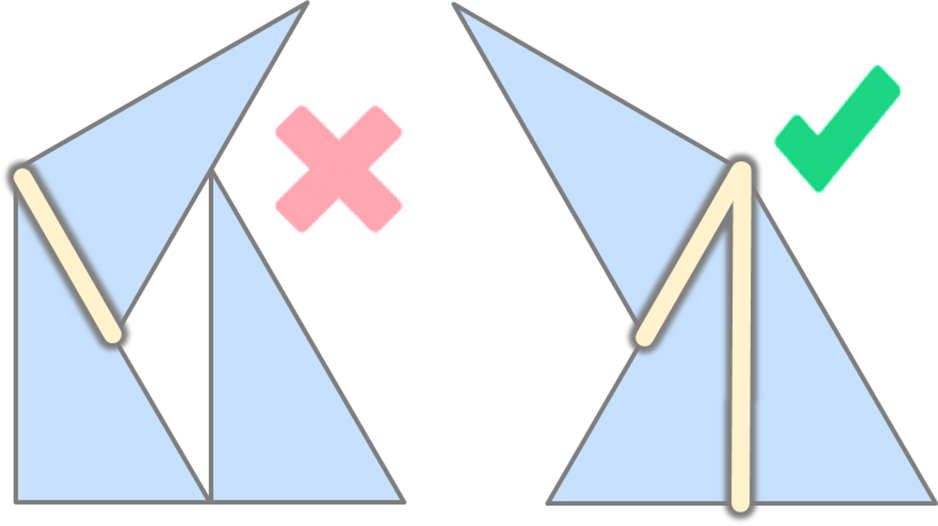}
\hspace{-30pt}
\vspace{-15pt}
\end{wrapfigure}
the total length of contacting edges between all adjacent nodes (see the right inset figure for illustrations); and 
(iii) we have a hard constraint that no two nodes are connected by edges in $\mathcal{E}_\text{ovl}$.

\section{TilinGNN for Tiling Generation}
\label{sec:method}

In this section, we first present the inputs we feed to TilinGNN (Section~\ref{ssec:nn_inputs}), the architecture of TilinGNN (Section~\ref{ssec:network_arch}), and the loss function we formulated for self-supervised training (Section~\ref{ssec:training_loss}).
Then, we present the procedure to employ a trained TilinGNN model to tile a given shape or region (Section~\ref{ssec:test}).

% XH's draft
%Our method contains four parts.
%First, to provide enough information for a network to select nodes to form tiling solutions, we enrich an adjacency graph by appending features to the %nodes and neighbor edges.
%Second, we present a novel graph convolutional network to solve our problem, which takes an adjacency graph with features on nodes and edges as input, and %output the probability of selecting each node in the graph.
%Third, we present the losses and how to train our network.
%Finally, we present the testing phase of our network, i.e., how to use the trained network to generate tiling solutions for an input $\mathcal{R}$.

%%%%%%%%%%%%%%%%%%%%%%%%%%%%%%%%%%%%%%%%%%%%%%%%%%%%%%%%%%%%%%%%%%%%%%

\subsection{Neural network inputs}
\label{ssec:nn_inputs}

In each iteration to train TilinGNN, we start by using a random 2D shape to crop and locate a subset of candidate tile locations \final{in} the ``superset'' of the target tile set $\mathcal{T}$; see again Figure~\ref{fig:overview}(b).
Note again that TilinGNN is trained per tile set.
Then, we construct an adjacency graph $\mathcal{G} = \{ \mathcal{V} , \mathcal{E}_\text{ovl}, \mathcal{E}_\text{nbr} \}$ to describe the connection and overlap relations between the cropped candidate tile locations, following the procedure presented in Section~\ref{ssec:enc_graph}.
Besides $\mathcal{G}$, we prepare the following two sets of inputs to train TilinGNN:
\begin{itemize}
\item
{\em Per-node vectors\/} $\{ \mathbf{v}_i \}$.
To start, we denote $N$ as the number of nodes in $\mathcal{V}$ and $N_t$ as the number of tile types in $\mathcal{T}$.
Also, we denote $A_i$ as the area of the tile ($i$-th candidate tile location) associated with the $i$-th node in $\mathcal{V}$ after normalized by the maximum tile area, i.e., $\max(A_i)$.
Then, we prepare an $(N_t$$+$$1)$-dimensional vector $\mathbf{v}_i$ per node in $\mathcal{V}$, where 
$i \in \{1,...,N\}$, 
the first element of $\mathbf{v}_i$ is $A_i \in [0,1]$ and
the remaining $N_t$ elements is a one-hot vector (a single \phil{`1'} with all the other \phil{'0's}) representing which of the $N_t$ tiles (i.e., tile type) in $\mathcal{T}$ associated with the $i$-th candidate tile location.
\item
{\em Per-neighbor-edge vectors\/} $\{ \mathbf{e}_j \}$.
We denote $N_e$ as the number of neighbor edges in $\mathcal{E}_\text{nbr}$ and $N_p$ as the number of all different relative poses
%(ignoring $n$-folded symmetry) 
between connectable tiles in $\mathcal{T}$, e.g., for the tile sets shown in \final{Figures~\ref{fig:super_graphs} (a) \& (b)}, $N_p$ are $8$ and \final{$32$}, respectively.
Also, we denote $L_{\max}$ as the maximum perimeter among the perimeters of all tile types in $\mathcal{T}$, and compute the length of the shared edge segment for each of the $N_p$ relative poses.
Then, we prepare an $(N_p$$+$$1)$-dimensional vector $\mathbf{e}_j$ per neighbor edge in $\mathcal{E}_\text{nbr}$, where 
$j \in \{1,...,N_e\}$,
the first element of $\mathbf{e}_j$ is the length of the shared edge segment associated with the $j$-th neighbor-edge connection in $\mathcal{E}_\text{nbr}$ after normalized by $L_{\max}$, 
and
the remaining $N_p$ elements is a one-hot vector representing which of the $N_p$ relative poses that the $j$-th neighbor edge associates with.
\end{itemize}
Note that we do not extract extra information for the overlap edges in $\mathcal{G}$, since overlap is a hard \final{constraint} and we must avoid all kinds of overlap connections.
Also, while there are many other geometric and topological information that we may include in $\{ \mathbf{v}_i \}$ and $\{ \mathbf{e}_j \}$, e.g., the coordinates of the tile location and orientation angle, we do not find them helpful in training TilinGNN.
%\phil{I assume that coordinates are really not helpful and please give one more example that you have explored before}

%%%%%%%%%%%%%%%%%%%%%%%%%%%%%%%%%%%%%%%%%%%%%%%%%%%%
\begin{figure}[t]
	\centering
	\includegraphics[width=0.99\linewidth]{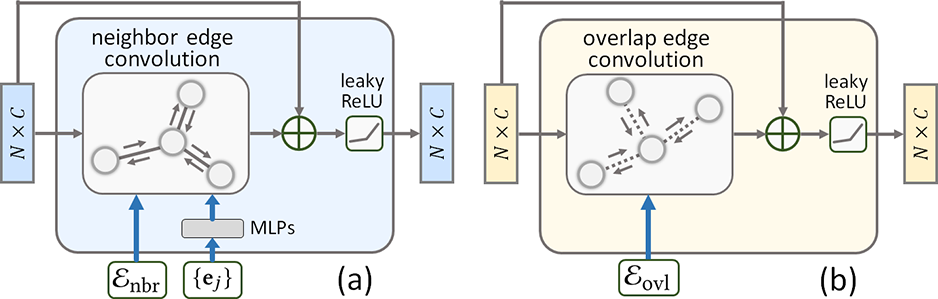}
	\ifdefined\negativevspace
	\vspace*{-1.5mm}
	\fi
	\caption{Illustration of the neighbor aggregation module (a) and the overlap aggregation module (b) inside TilinGNN shown in Figure~\ref{fig:net_architec}.}
	\label{fig:graph_conv}
	\ifdefined\negativevspace
	\vspace*{-1mm}
	\fi
\end{figure}
\begin{figure*}[t]
	\centering
	\includegraphics[width=0.99\linewidth]{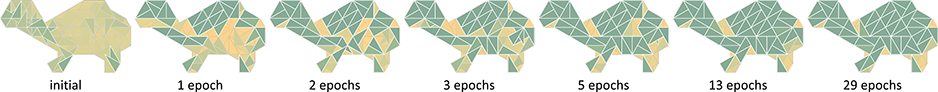}
	\ifdefined\negativevspace
	\vspace*{-1.5mm}
	\fi
	\caption{Visualizing the network outputs when using TilinGNNs trained for different number of epochs to test on the \textsc{Tortoise} shape.
	Note that we render the candidate tile placements from low (yellow) to high (green) network-output probabilities, so those with high probabilities are less occluded.
	From the results, we can see that TilinGNN can gradually learn over the training process to give higher probabilities to select non-overlapping and connecting tiles.}
%
%	\caption{Visualizing the selection probabilities of the test shape \textsc{tortoise} predicted by models trained after different number of epochs. 
%		We map the probability of selecting each tile to a color, where the redder means higher selection probability.}
	\label{fig:vis_probs}
	\ifdefined\negativevspace
	\vspace*{-1mm}
	\fi
\end{figure*}
%%%%%%%%%%%%%%%%%%%%%%%%%%%%%%%%%%%%%%%%%%%%%%%%%%%%

\subsection{Network Architecture}
\label{ssec:network_arch}

Figure~\ref{fig:net_architec} shows the overall architecture of TilinGNN, whereas Figure~\ref{fig:graph_conv} shows the detailed structure of two major modules inside TilinGNN, i.e., the neighbor aggregation and overlap aggregation modules.
Overall, TilinGNN is a {\em two-branch graph convolutional neural network\/} that progressively propagates and aggregates node features over the neighbor edges and overlap edges.
Here, we denote $\mathbf{F}^l$ and $\mathbf{G}^l$ as the node features for the two modules, where $l$ is a nonnegative integer that denotes layer.
The inputs to TilinGNN include an adjacency graph $\mathcal{G}$ and the associated per-node and \final{per-edge} information $\{ \mathbf{v}_i \}$ and $\{ \mathbf{e}_j \}$, whereas its output is a vector of $N$ values, indicating the probability of selecting each node \final{in $\mathcal{G}$}, i.e., a candidate tile location, for the tiling generation.

From left to right in the network architecture diagram shown in Figure~\ref{fig:net_architec}, TilinGNN first uses a set of shared multi-layer \final{perceptrons} (MLPs) to process $\{ \mathbf{v}_i \}$ and generate node feature $\mathbf{F}^0$ (dimension: $N \times C$), which is a set of $N$ per-node feature vectors, each of $C$ channels.
Then, we pass $\mathbf{F}^0$ independently into a neighbor aggregation module and an overlap aggregation module, which form the first feature aggregation layer.
The output of the neighbor aggregation module is then aggregated (multiplied) with the output of the overlap aggregation module, which is $\mathbf{G}^0$ (dimension: $N \times C$), to produce node feature $\mathbf{F}^1$.
In this way, we can propagate each node feature vector in $\mathbf{F}^0$ through the node's associated neighbor edge(s) and overlap edge(s) in the adjacency graph for one step.

To enable TilinGNN to learn effective node features, we should further propagate the node features for more steps over the graph, which is essentially the spatial domain for tiling.
Hence, we further feed $\mathbf{F}^1$ into the second feature aggregation layer, etc., and subsequently generate $\mathbf{F}^2$ up to $\mathbf{F}^L$, where $L$ is the number of feature aggregation layers in TilinGNN.
In the end, we produce the overall node feature $\mathcal{F}$ (dimension: $N \times C \times (L+1)$) by concatenating all the node features $\{ \mathbf{F}^0 , \mathbf{F}^1 , ..., \mathbf{F}^L \}$.
We then feed $\mathcal{F}$ into a set of shared MLPs with sigmoid activation to map each node feature vector in $\mathcal{F}$ to predict a probability value.
Note also that we set $L = 20$ and $C = 64$; see the implementation details in Section~\ref{sec:results}.

%%%%%%%%%%%%%%%%%%%%%%%%%%%%%%%%%%%%%%%%%%%%%%
\ifdefined\negativevspace
\vspace*{-3pt}
\fi
\paragraph{Neighbor aggregation module}
This module focuses on aggregating node features over neighbor edges $\mathcal{E}_\text{nbr}$; see Figure~\ref{fig:graph_conv}(a).
\final{We denote} $\mathbf{F}^l = \{ \mathbf{f}^l_1 , \mathbf{f}^l_2 , ... , \mathbf{f}^l_N \}$ as the node features in the $l$-th layer.
From $\mathcal{E}_\text{nbr}$, $\mathbf{F}^l$, and $\{ \mathbf{e}_j \}$, the module first uses a set of shared MLPs (for $l$-layer) to process $\{ \mathbf{e}_j \}$ and generate $\Phi^l(\mathbf{e}_j)$ per edge in $\mathcal{E}_\text{nbr}$.
The dimension of $\Phi^l(\mathbf{e}_j)$ is $C \times C$.
Then, we employ an edge-conditioned convolution layer~\cite{simonovsky2017dynamic} to aggregate features from the neighboring nodes (along neighbor edges) conditioned on the associated edge features for each node\final{, with leaky ReLU (denoted as LReLU) as the activation output function}:
\begin{equation}
\label{eq:nbr_conv}
\mathbf{f}^{l+1}_{i}
=
\mathrm{LReLU}\big(
\mathbf{f}^l_i \cdot W^l
+
\sum_{\{k,j\} \in N_\text{nbr}(i)} \mathbf{f}^l_k \cdot \Phi^l(\mathbf{e}_j) \big) \ ,
\end{equation}
where
$W^l$ is learnable weight (dimension: $C \times C$) and
$N_\text{nbr}(i)$ is the set of indices of nodes ($k$) and edges ($j$) connected with the $i$-th node in $\mathcal{V}$ via neighbor edges.
Also, we add residual connections in the module (see again Figure~\ref{fig:graph_conv}(a)) to avoid the problem of vanishing gradient\final{~\cite{he2016deep}}, since the network is deep.

%%%%%%%%%%%%%%%%%%%%%%%%%%%%%%%%%%%%%%%%%%%%%%
\ifdefined\negativevspace
\vspace*{-3pt}
\fi
\paragraph{Overlap aggregation module}
Similar to the neighbor aggregation module, the overlap aggregation module also takes \final{the} node feature from the previous layer as input.
Typically, its input node feature can be $\mathbf{F}^0$ (for 1st layer) or $\mathbf{G}^l$ (for 2nd to $L$-th layers); see Figure~\ref{fig:net_architec}.
For simplicity, we denote it as $\mathbf{G}^l = \{ \mathbf{g}^l_1 , \mathbf{g}^l_2 , ..., \mathbf{g}^l_N \}$.
Then, we employ the convolution layer proposed by~\cite{xu2019powerful} to perform convolutions on the overlap edges for each node:
\begin{equation}
\label{eq:ovl_conv}
\mathbf{g}^{l+1}_{i}
=
\mathrm{LReLU}\big(
\Theta^l \big( (1+\epsilon\final{^l})\mathbf{g}^l_i + \sum_{k \in N_\text{ovl}(i)} \mathbf{g}^l_k \big) \big),
\end{equation}
where
$\epsilon\final{^l}$ is a learnable value (initialized as zero and modified adaptively by the network optimizer);
$N_\text{ovl}(i)$ is the set of indices of nodes ($k$) connected with the $i$-th node in $\mathcal{V}$ via overlap edges; and
$\Theta^l(\cdot)$ is a differentiable function implemented by a set of shared MLPs \final{using also leaky ReLU as the activation function}.
Note that we use a different mechanism to aggregate overlap-related node features, since \cite{xu2019powerful} has been shown to be effective for scenarios, in which the graph has no edge labels.

\ifdefined\negativevspace
\vspace*{-3pt}
\fi
\paragraph{Discussion}
TilinGNN can also be viewed as a binary classification network on nodes in the adjacency graph, such that nodes predicted with high (or low) probability are likely (or unlikely) belonging to the final tiling.
However, unlike general classification problems, we have a large amount and different types of edge connections in our graph, and the selected high-probability nodes should not locate next to one another, due to the overlap constraint.
Hence, we cannot directly employ existing classification networks to our problem.

% XH's draft:
%\begin{itemize}
%	\item Our problem is a node classification task, just like those segmentation tasks. But in those segmentation tasks, the nodes in the same class %are typically close to each other.
%	In our problem, the selected nodes tend to be seperated to avoid collision.
%	\item We tried to employ a pooling layer to summarize a global feature, and concatenate with the final output feature. However, we found no obvious %improvements in the results. 
%\end{itemize}
%
%
%
%
%\phil{move these to Sec 6:}
%
%\paragraph{Discussion}
%\begin{itemize}
%	\item Our problem is a node classification task, just like those segmentation tasks. But in those segmentation tasks, the nodes in the same class %are typically close to each other.
%	In our problem, the selected nodes tend to be seperated to avoid collision.
%	\item We tried to employ a pooling layer to summarize a global feature, and concatenate with the final output feature. However, we found no obvious %improvements in the results. 
%\end{itemize}
%
%\paragraph{Hyperparameters.}
%Major parameters in our network:
%\begin{itemize}
%	\item 
%	The depth of our network, $L$, which we set as $20$ in our implementation, which is large enough for each node to receive message from any other %nodes.
%	\item 
%	the width of our network, which is set to be 64.
%\end{itemize}

%%%%%%%%%%%%%%%%%%%%%%%%%%%%%%%%%%%%%%%%%%%%%%%%%%%%%%%%%%%%%%%%%%%%%%

\subsection{Loss Function}
\label{ssec:training_loss}

We denote $x_i \in [0,1]$ as the network-output probability \final{of} selecting the $i$-th node (candidate tile location) in $\mathcal{G}$.
\final{Similar to~\cite{leimkuhler-2019-deep-point}, we formulate loss terms on $x_i$ to train TilinGNN in a self-supervised manner}, where $w_\text{a}$, $w_\text{o}$, and $w_\text{e}$ below are weights:
\begin{itemize}
\item 
To maximize the tiling coverage of the target region, we make use of the normalized tile area $A_i$ (see Section~\ref{ssec:nn_inputs}) to define
$$
L_\text{a} \ = \ 1 - w_\text{a} \log_e(\frac{\sum_i A_i x_i}{\sum_i A_i}) \ .
$$
\item
To minimize the tile overlaps, we define
$$
L_\text{o} \ = \ 1 - w_\text{o} \cdot \frac{1}{\left|\mathcal{E}_\text{ovl}\right|}
\sum_{\{i,k\} \in \mathcal{E_\text{ovl}}} \log_e(1 - x_i x_k) \ .
$$
\item 
To avoid holes among the selected tile locations, we maximize the total length of the contacting edge segments between the nodes connected by neighbor edges, and define
$$
L_\text{e} \ = \ 1 - w_\text{e} \cdot \frac{1}{\left|\mathcal{E}_\text{nbr}\right|} 
\sum_{\{i,k\} \in \mathcal{E_\text{nbr}}} \log_e( \frac{x_i x_k L_{i,k}}{L_{\max}} ) \ ,
$$
where $L_{i,k}$ is the length of the contacting edge segment between the $i$-th and $k$-th nodes, and $L_{\max}$ is the maximum tile perimeter, as defined earlier in Section~\ref{ssec:nn_inputs}.
\end{itemize}

Then, we formulate the overall loss \final{as} a product of $L_\text{a}$, $L_\text{o}$, and $L_\text{e}$, and aim to minimize this overall loss in the network training.
In our initial attempt, we use a weighted sum to combine the three terms, but the network training does not converge and yields poor results.
Our log-based design is inspired by~\cite{toenshoff-2019-run}, in which they formulate losses for constraint satisfaction problems.
Note also that the range of each term is from one to $+\infty$, due to the use of logarithms in the formulations of the terms.
Hence, setting a larger weight on a term will increase its strength.
In practice, we set $w_\text{o}$, $w_\text{a}$, and $w_\text{e}$ as 10.0, 1.0, and 0.02, respectively, since $L_\text{o}$ has the top priority to help avoid tile overlaps, whereas $L_\text{a}$ has the second priority as being the major optimization objective.

Figure~\ref{fig:vis_probs} shows the visualizations of the network outputs when using TilinGNNs trained for different numbers of epochs to test on the same shape.
To produce each visualization, we sort all candidate tile locations by their network-output probabilities and render them from low to high probability with color coding, so tile locations among the highest probabilities (green) are less occluded than those with low probabilities (yellow).
From Figure~\ref{fig:vis_probs}, we can see that the first few visualizations are more chaotic, whereas the chosen tile locations are better revealed in the latter visualizations.

\subsection{Tiling Generation}
\label{ssec:test}

Given target shape $\mathcal{R}$ and tile set $\mathcal{T}$, we take the following two steps to generate a tiling on $\mathcal{R}$ using the TilinGNN trained on $\mathcal{T}$.

%%%%%%%%%%%%%%%%%%%%%%%%%%%%%%%%%%%%%%%%%%%%%%
\ifdefined\negativevspace
\vspace*{-3pt}
\fi
\paragraph{Step one: find candidate tile locations for $\mathcal{R}$}
First, we analyze $\mathcal{T}$'s tiling pattern (e.g., see the superset in Figure~\ref{fig:overview}(a)) to find
(i) the minimum angle $\theta$ to rotate the pattern, and
(ii) the minimum translation $\Delta x$ (and $\Delta y$) in $X$ (and $Y$) dimension to shift the pattern,
such that the pattern aligns itself by rotation and translation symmetry.

Given $\mathcal{R}$ and a scale factor on $\mathcal{R}$, there are still many ways (positions and orientations) of putting $\mathcal{R}$ over $\mathcal{T}$'s superset.
So, we randomly sample six rotation angles in $[0,\theta)$ and nine translation vectors in $[0, \Delta x)$ and $[0,\Delta y)$, and use 54 combinations of these samples to put $\mathcal{R}$ on $\mathcal{T}$'s superset and crop candidate tile locations.
We then pick the top $K$ configurations that lead to the largest total tile area, and take the set of candidate tile locations of each configuration as $\mathcal{P}$, an input to step two; see the leftmost side of Figure~\ref{fig:overview}(c) for an example.
In our implementation, we set $K$ = 4 to 20, depending on the problem complexity.
Note that we may skip step one, if we use our interactive design interface (see Section~\ref{ssec:results}).

% XH's draft
%\begin{itemize}
%	\item First, we analyze the patterns in $\mathcal{P}$. \\
%	We analyze the minimum rotation angle $\theta$ that can bring $\mathcal{P}$ to coincident with itself (k-fold symmetry).\\
%	We analyze the minimum translation vector $\vec{t} = (x, y)$ that can bring $\mathcal{P}$ to coincident with itself . \\
%	\item Then, we randomly sample $6$ rotation angles in $[0, \theta]$, and sample $20$ vectors in the 2D space of ($[0, x]$, $[0,y]$).
%	\item parallel
%\end{itemize}

%%%%%%%%%%%%%%%%%%%%%%%%%%%%%%%%%%%%%%
% pseudo-code

\begin{algorithm}[!t] % enter the algorithm environment
	\SetAlgoNoLine
	\KwData{Target shape $\mathcal{R}$ and TilinGNN trained for target tile set $\mathcal{T}$}
%	\KwData{Target shape $\mathcal{R}$; TilinGNN trained for target tile set $\mathcal{T}$; and a set of candidate tile placements $\mathcal{P}$}
%	\KwData{Adjacency graph $\mathcal{G}$ with associated $\{\mathbf{v}_i\}$ and $\{\mathbf{e}_j\}$, and trained TilinGNN for $\mathcal{T}$}
	\KwResult{a set of tiles in $\mathcal{R}$}
	%
	% Step 1
	\final{$\mathcal{S}$
	\hspace*{-0.45mm} $\gets \emptyset$
	\hspace*{5mm} \tcp{\footnotesize final tiling solution}}
	$\{ \mathcal{P}_1 , \mathcal{P}_2 , ..., \mathcal{P}_K \}$
	\\
	\hspace*{5mm}
	$\gets$ Find sets of candidate tile placements of top $K$ areas for $\mathcal{R}$
	\\
	\For{$\mathcal{P} \leftarrow \mathcal{P}_1$ \KwTo $\mathcal{P}_K$}{
	\vspace*{0.5mm}
	$\mathcal{G}$
	\hspace*{1.12mm} $\gets$ Create adjacency graph, $\{\mathbf{v}_i\}$, and $\{\mathbf{e}_j\}$ \final{from} $\mathcal{P}$
	\\
	\final{$\mathcal{S}_\mathcal{P}$
	\hspace*{-0.45mm} $\gets$ $\emptyset$
	\hspace*{16.5mm} \tcp{\footnotesize tiling solution using $\mathcal{P}$}}
	$k$
	\hspace*{1.9mm} $\gets 1$
	\hspace*{16.75mm} \tcp{\footnotesize number of rounds}
	$\mathbf{p}$
	\hspace*{2.0mm} $\gets \{1,...,1\}$
	\hspace*{8mm} \tcp{\footnotesize tile selection probability} 
	%
	% Step 1
	\While{$\mathcal{G} \neq \emptyset$}{
		\vspace*{0.5mm}
		$\mathbf{x}$
		\hspace*{4.15mm} $\gets$ Apply TilinGNN to test $\mathcal{G}$
		\\
		$\mathbf{p}_i$
		\hspace*{3.05mm} $\gets$ $(\mathbf{p}_i^{k-1} \ \cdot \ \mathbf{x}_i)^\frac{1}{k}$, $\forall \mathcal{V}_i \in \mathcal{G}$
		\\
		\vspace*{0.5mm}
		$\mathbf{p}^\text{temp} \gets$ sort $\mathbf{p}$ in descending order
		\\
		\vspace*{1mm}
		\For{$i\leftarrow 1$ \KwTo $|\mathbf{p}^\text{temp}|$}{
			$j \gets$ get index to node in $\mathcal{G}$ for $\mathbf{p}^\text{temp}_i$
			\\
			%$p_0 \gets$ pop $\mathbf{p}^\text{temp}_1$
			%	\hspace*{15mm} \tcp{\footnotesize pop \& remove 1st element}
			%
			\uIf{tile of node $j$ overlaps with any tile in \final{$\mathcal{S}_\mathcal{P}$}}{
				break
				\hspace*{22.2mm} \tcp{\footnotesize exit inner loop}
			}
			\uIf{$e^{(\mathbf{p}^\text{temp}_i-1)} >$ random value in $[0,1]$}{
				\final{$\mathcal{S}_\mathcal{P}$} $\gets$ add tile of node $j$
				\hspace*{0.5mm} \tcp{\footnotesize select by prob.}
			}
		}
		$\{ \mathcal{G}, \mathbf{p} \} \gets$ remove all nodes in $\mathcal{G}$ that overlap with \final{$\mathcal{S}_\mathcal{P}$}
		\\
		$k$ $\gets$ $k+1$
	}
	\final{$\mathcal{S}$ $\gets$ select the one with lower loss from $\mathcal{S}$ and $\mathcal{S}_\mathcal{P}$}
	}
	\textbf{output} $\mathcal{S}$
	\caption{The Overall Tiling Procedure}
	\label{alg:algo_tiling}
\end{algorithm}

% XH's draft
%
%\begin{algorithm}[!t] % enter the algorithm environment
%	\SetAlgoNoLine
%	\KwData{Candidate tile graph $\mathcal{G}$} 
%	\KwResult{set of indexes of the selected nodes $\mathcal{S}$.}
%	$\mathcal{S} \gets \emptyset$, $\mathcal{G}' \gets \mathcal{G}$, $k \gets 1$ \\
%	$\mathbf{p} \gets [1,...,1]$     \tcp{\footnotesize the accumulated selection probability} 
%	\While{$\mathcal{G}'$ is not empty}{
%		$\mathbf{x}$ $\gets$ $f(\mathcal{G}')$ \tcp{\footnotesize network predict}
%		$\mathbf{p}[i]$ $\gets$ $(\mathbf{p}[i]^{k-1} \ \times \ \mathbf{x}[i])^\frac{1}{k}$, $\forall i \in \{0,...,|\mathbf{p}|\}$ \\
%		\vspace*{0.5mm}
%		$\mathbf{p}^\text{temp} \gets \mathbf{p}$ \\
%		\While{$\left| \mathbf{p}^\text{temp} \right| > 0 $}{ 
%			j $\gets$ index\_sample\_and\_pop($\mathbf{p}^\text{temp}$) \\
%			\uIf{node $j$ is already labeled}{
%				break    \tcp{\footnotesize node $j$ overlap with selected nodes}
%			}
%			\Else{
%				append $j$ to $\mathcal{S}$ \\
%				label the overlap neighbors of node $j$ \\
%			}
%		}
%		$\mathcal{G}'$, $\mathbf{p}$ $\gets$ remove\_labeled\_nodes($\mathcal{G}'$, $\mathbf{p}$) \\
%		$k$ $\gets$ $k+1$ \\
%	}
%	\textbf{return} $\mathcal{S}$
%	\caption{The probabilistic greedy algorithm.}
%	\label{alg:search_code}
%\end{algorithm}

%%%%%%%%%%%%%%%%%%%%%%%%%%%%%%%%%%%%%%%%%%%%%%%%%%%%
\begin{figure}[t]
	\centering
	\includegraphics[width=0.99\linewidth]{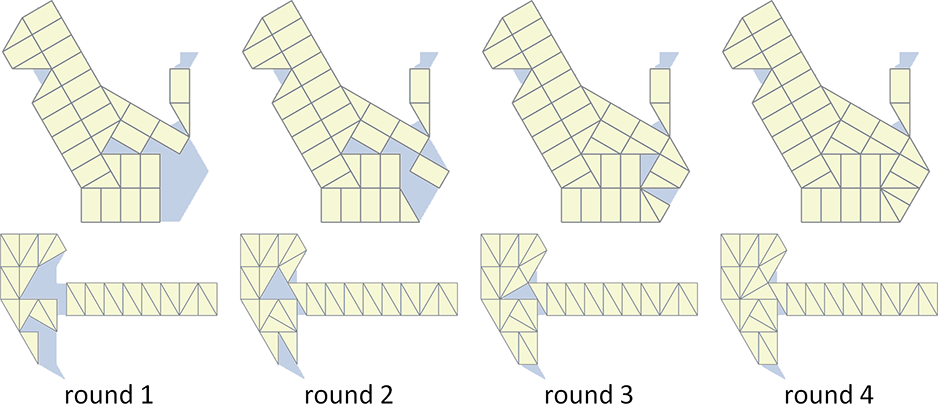}
	\ifdefined\negativevspace
	\vspace*{-1.5mm}
	\fi
	\caption{Running examples of using Algorithm~\ref{alg:algo_tiling} to tile the {\sc Cat} and {\sc Hammer} shapes.
	Note that the tiling process usually finishes in one/two rounds; we intentionally pick these two cases \final{with more} rounds for illustration.
	\final{Also, we depict the union of all candidate tile locations in blue under the tiling solutions as a means to visualize the holes and gaps in the results.}
	}
	\label{fig:running_example}
	\ifdefined\negativevspace
	\vspace*{-1mm}
	\fi
\end{figure}
%%%%%%%%%%%%%%%%%%%%%%%%%%%%%%%%%%%%%%%%%%%%%%%%%%%%

%%%%%%%%%%%%%%%%%%%%%%%%%%%%%%%%%%%%%%%%%%%%%%
\ifdefined\negativevspace
\vspace*{-3pt}
\fi
\paragraph{Step two: tiling using a trained TilinGNN model}
From each $\mathcal{P}$ obtained in step one, we construct \final{adjacency} graph $\mathcal{G}$ and its associated network inputs $\{ \mathbf{v}_i \}$ and $\{ \mathbf{e}_j \}$, and apply the TilinGNN trained on $\mathcal{T}$ to produce a probability vector on the candidate tile locations in $\mathcal{P}$.
Next, we sort the candidate tile locations by the probabilities, and select them in descending order of the probability values.
If the next selected tile conflicts (overlaps) with any chosen tile, we stop the selection, create a new (partial) \final{adjacency} graph for the remaining candidate tile locations, and start another round of tiling with the trained TilinGNN on the new graph; see Figure~\ref{fig:overview}(c).

Algorithm~\ref{alg:algo_tiling} gives the overall tiling procedure.
Very different from the traditional search tools, its running time is roughly linear to the number of nodes in $\mathcal{G}$, so it usually finishes in less than a minute in our experiments.
Particularly, the inner loop adopts a probabilistic selection mechanism inspired by the acceptance probability model in general simulated annealing methods~\cite{cagan1998simulated} to accept tile candidates, so Algorithm~\ref{alg:algo_tiling} has the chance of producing different tiling solutions in different runs.
This \final{strategy} helps \final{to} increase the tiling diversity and avoid holes in the results.
Note also that since the network output is already of good quality (see the rightmost result in Figure~\ref{fig:vis_probs}), we usually need only a few (one to two) rounds of tiling in Algorithm~\ref{alg:algo_tiling}; see Figure~\ref{fig:running_example} for two \final{typical results with more steps}.
\final{Note that when we present our tiling solutions (started from Figure 12), we show the 2D region covered by the union of all candidate tile locations in blue color under the tiling solutions for better visualization of the holes and gaps in the results.}

\paragraph{\final{Discussion}}
\final{At the beginning of this research, we attempted to achieve end-to-end training by trying to formulate the loss on Algorithm~\ref{alg:algo_tiling}'s output.
%by computing the loss values directly using the binarized outputs.
However, since we cannot directly compute the gradients on such binarized outputs, we thus make use of Algorithm~\ref{alg:algo_tiling} to interpret the outputs of TilinGNN instead.}

%%%%%%%%%%%%%%%%%%%%%%%%%%%%%%%%%%%%%%

%%%%%%%%%%%%%%%%%%%%%%%%%%%%%%%%%%%%%%%%%%%%%%%%%%%%
\begin{figure}[t]
	\centering
	\includegraphics[width=0.92\linewidth]{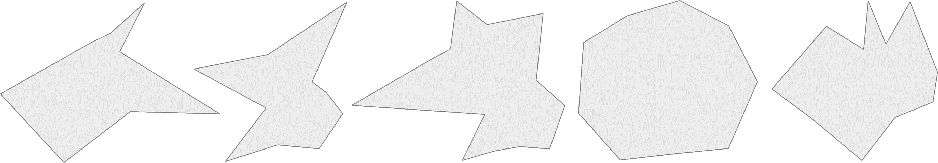}
     \vspace*{-1mm}
	\caption{\final{Random shapes (randomly picked from the 12,000 random shape set) we employed for cropping the superset of the candidate tile locations.}}
	\label{fig:rand_shapes}
\end{figure}

\section{Results and Experiments}
\label{sec:results}

%%%%%%%%%%%%%%%%%%%%%%%%%%%%%%%%%%%%%%%%%%%%%%%%%%%%%%%%%%%%%%%%%%%%%%%%%%%%%%%%%%

\subsection{Implementation Details}
\final{We implemented TilinGNN in Python 3.7 using PyTorch~\cite{NEURIPS2019-pytorch}.}
We employed Numpy~\cite{numpy} to manipulate the arrays and their computations, and Shapely~\cite{shapely} for geometric computations such as collision detection.

%%%%%%%%%%%%%%%%%%%%%%%%%%%%%%%%%%%%%%%%%
\ifdefined\negativevspace
\vspace*{-3pt}
\fi
\paragraph{Training data preparation.}
For each tile set, we pre-computed its superset, i.e., a set of candidate tile locations.
%(see Figure~\ref{fig:overview}(a)).
Table~\ref{fig:statistics_tileset} reports the tile set statistics, in which we sort the table rows by the tile set complexity estimated by the mean degree of graph nodes, i.e., $\frac{2(\left|\mathcal{E}_\text{ovl}\right| + \left|\mathcal{E}_\text{nbr}\right|)}{ \left|\mathcal{V}\right|}$.
%, and sorted the rows in the table by this quantity.
%
%\xh{To compare the complexity of different tile sets, we first scale the tiles in the tile sets, such that the highest tile in every tile set has the same height.
%Then, we crop the candidate tile locations of different tile sets using the circle with the same size, in such a way that the union area of the cropped tiles of different tile sets have almost the same area.}
%\phil{doesn't seem to be from the statistics, but complete the paper first and come back, if you have time}
%
On the other hand, we prepared 12,000 \final{random shapes for cropping the superset (see Figure~\ref{fig:overview}(b) for how these shapes are used)}, and randomly divided them into a training set (10,000) and a validation set (2,000).
%in the network training process.
So, one epoch in network training takes 10,000 iterations with the shapes in the training set.
To produce a random shape, we randomized its number of vertices in range $\{3, ..., 20\}$ and its size in range $[0.3,0.8]$, then constructed the shape by randomizing its vertex coordinates.
Further, we discarded and re-generated a shape, if it contains any self-intersection.
\final{Figure~\ref{fig:rand_shapes} shows some of the random shapes (randomly picked from the 12,000 set), showing that our training process considers both convex and concave shapes.}

% XH's draft:
%\vspace*{-3pt}
%\paragraph{Processing a tile set.}
%Given a tile set, we first pre-compute the candidate tile placements $\mathcal{P}$, following the method in section~\ref{ssec:tile_set}.
%To train the network, we generated a synthetic dataset of random shapes.
%To generate each random shape, we first generate a random integer number $K$ in $\{3, ..., 20\}$ as the number of vertice of the shape, and a floating %number $d\in[0.1]$ as the scale of the shape.
%After that, we uniformly sample $K$ points inside a circle of diameter of $d \cdot L_\mathcal{P}$, where $L_\mathcal{P}$ is the diameter of $\mathcal{P}$.
%Finally, we connect these point counter-clockwise around the center of the circle to form a polygon, and convert the input shape into an adjacency graph %accordingly. 
%For each tile set, we created 10,000 shapes for training and 2000 shapes for the evaluation.
%Table~\ref{fig:tile_set} shows the statistics of different tile sets.
%We use the average degree of the adjacency graph $\frac{2(\left|\mathcal{E}_\text{ovl}\right| + \left|\mathcal{E}_\text{nbr}\right|)}{ %\left|\mathcal{V}\right|}$, to quantify the complexity of a data set.

\begin{table}[t]
	\centering
	\caption{Tile set statistics.
		From left to right, for each tile set, we show
		its tile types,
		number of candidate tile locations in its superset, %$\left|\mathcal{P}\right|$, 
		number of overlap edges ($\left|\mathcal{E}_\text{ovl}\right|$), 
		number of neighbor edges ($\left|\mathcal{E}_\text{nbr}\right|$), and 
		mean degree of graph nodes ($\frac{2(\left|\mathcal{E}_\text{ovl}\right| + \left|\mathcal{E}_\text{nbr}\right|)}{ \left|\mathcal{V}\right|}$) in its adjacency graph,
		batch size in network training,
		training \final{time}, 
		and test time \final{(for 20 runs)}.
	}
%	\caption{Statistics of the tile sets in our experiments.
%		From left to right, we show the tiles in the tile set, the number of possible tile placements $\left|\mathcal{P}\right|$, 
%		the number of overlap edges in the adjacency graph $\left|\mathcal{E}_\text{ovl}\right|$, 
%		the number of neighbor edges in the adjacency graph $\left|\mathcal{E}_\text{nbr}\right|$, 
%		the average degree of the adjacency graph, calculated by $\frac{2(\left|\mathcal{E}_\text{ovl}\right| + \left|\mathcal{E}_\text{nbr}\right|)}{ \left|\mathcal{V}\right|}$,
%		the batch size of in the training,
%		the training time for the corresponding tile set, and
%		the time for testing on the same set of tiling regions.
%	}
	\ifdefined\negativevspace
	\vspace*{-1mm}
	\fi
	\includegraphics[width=0.99\linewidth]{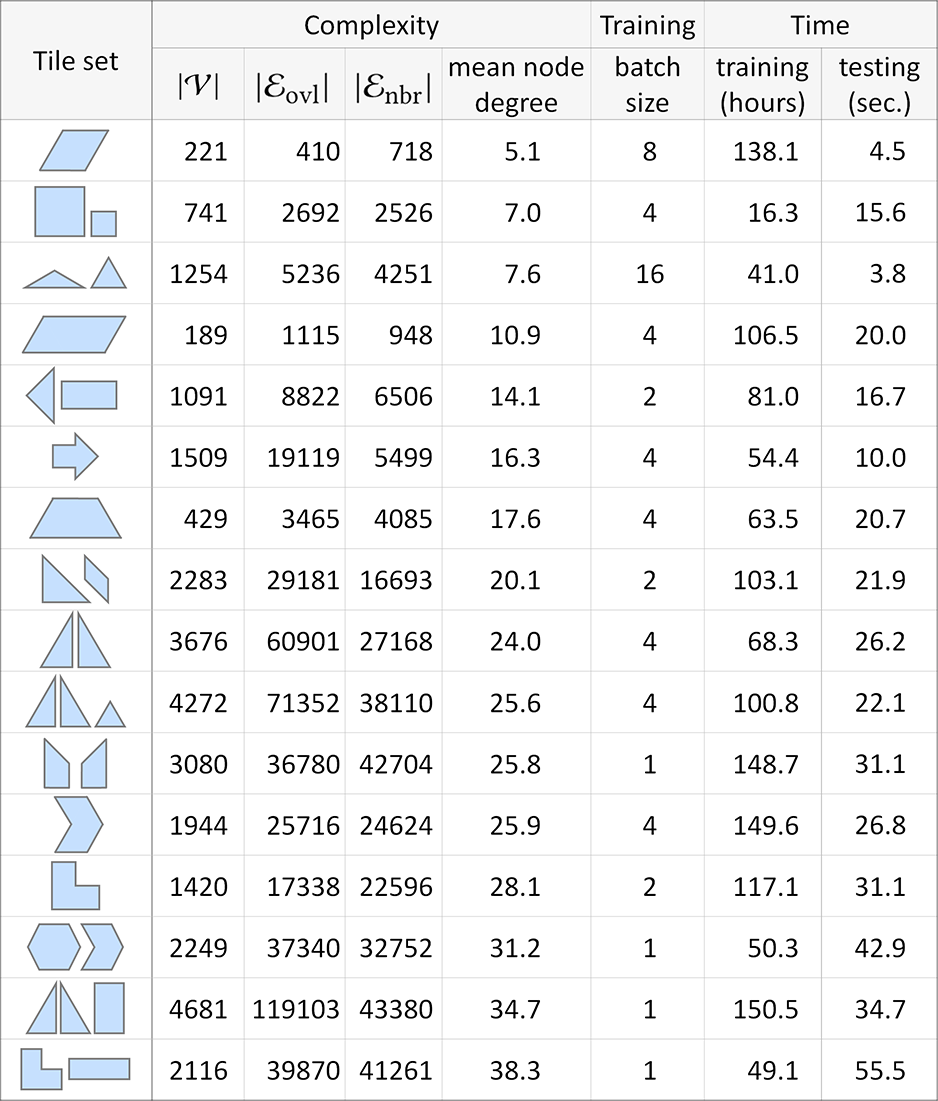}
	\label{fig:statistics_tileset}
	%\vspace*{-1mm}
\end{table}

%%%%%%%%%%%%%%%%%%%%%%%%%%%%%%%%%%%%%%%%%%%%%%%%%%%%
\begin{figure*}[t]
	\centering
	\includegraphics[width=0.97\linewidth]{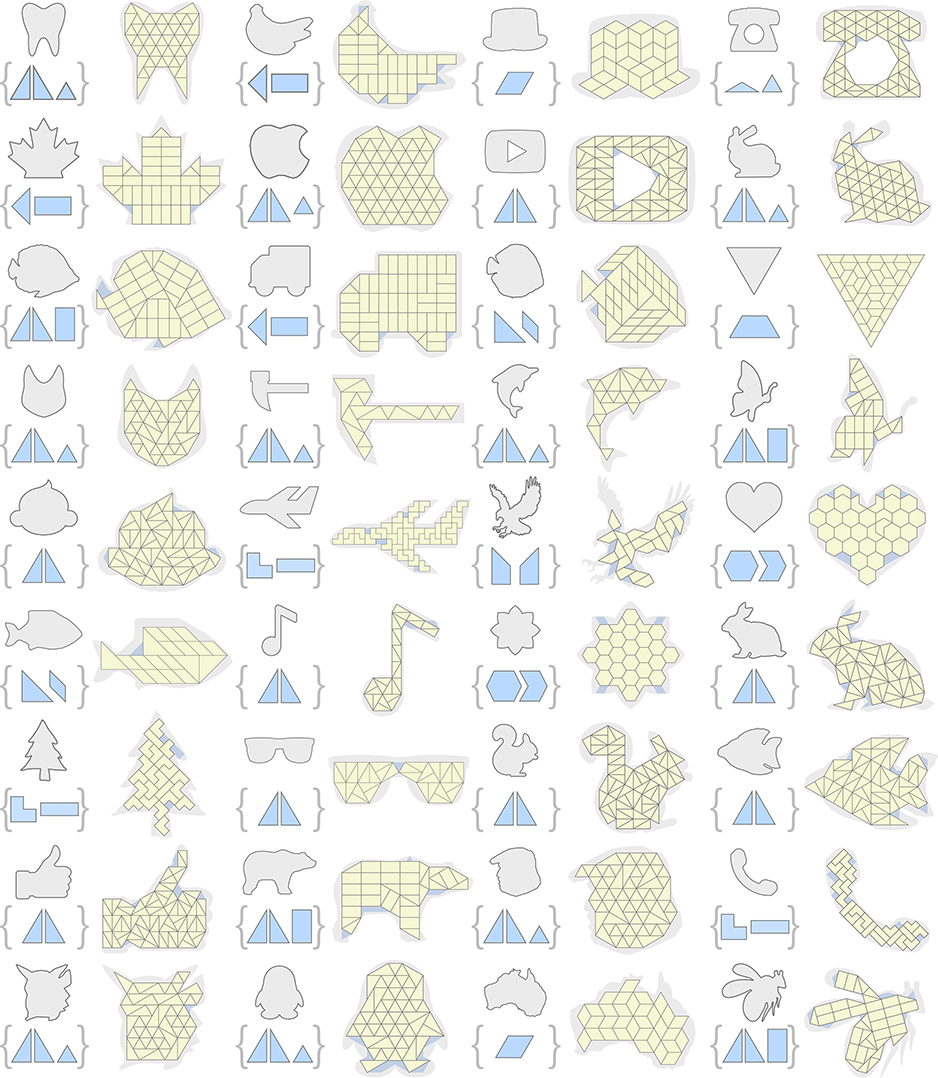}
	\vspace*{-1.75mm}
	\caption{A gallery, showcasing the tiling solutions produced by our learn-to-tile approach on 36 different shapes. \final{Under each tiling solution, we show the original input shape in gray and the region covered by the union of all the candidate tile locations (inside the input shape) in blue, as references.}}
	\label{fig:gallery}
	%\vspace*{-1mm}
\end{figure*}
%%%%%%%%%%%%%%%%%%%%%%%%%%%%%%%%%%%%%%%%%%%%%%%%%%%%

\begin{figure*}[t]
	\centering
	\includegraphics[width=0.97\linewidth]{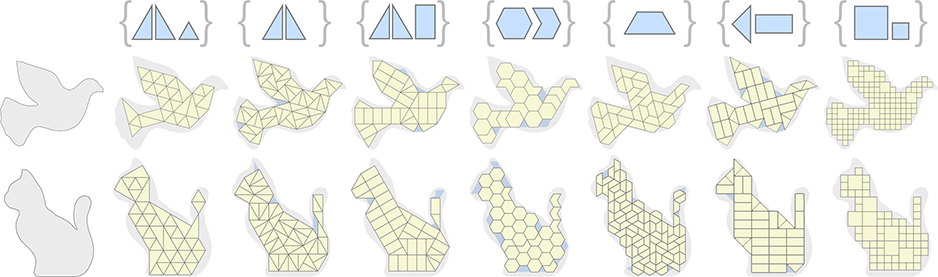}
	\ifdefined\negativevspace
	\vspace*{-1mm}
	\fi
	\vspace*{-1.5mm}
	\caption{Using our approach to tile the same shape ({\sc Dove} and {\sc Cat}) using different tile sets.}
	\label{fig:shape_with_diff_tiles}
\end{figure*}

%%%%%%%%%%%%%%%%%%%%%%%%%%%%%%%%%%%%%%%%%
\ifdefined\negativevspace
\vspace*{-3pt}
\fi
\paragraph{Network training.}
We trained the TilinGNN model on an NVidia Titan Xp GPU using the Adam optimizer~\cite{kingma2014adam} with learning rate $10^{-3}$, and ended the training, if the loss stopped to reduce.
Overall, it took one to five days to train a model for 20 to 80 epochs.
See Table~\ref{fig:statistics_tileset} for the training time and batch size.

% XH's draft:
%\vspace*{-3pt}
%We train each model on an NVidia Titan Xp GPU.
%We use Adam optimizer~\cite{kingma2014adam} with  learning rate $10^{-3}$.
%The training of different tile sets often take 20 to 80 epochs, depending on one the loss decreasing.
%See Table~\ref{fig:tile_set} again for the training time and batch size for all tile sets in our experiments.
%It often takes one to five days to train a network, where we stop our network when we do not see any significant loss decrease.

%%%%%%%%%%%%%%%%%%%%%%%%%%%%%%%%%%%%%%%%%%%%%%%%%%%%
\begin{figure}[t]
	\centering
	\includegraphics[width=0.99\linewidth]{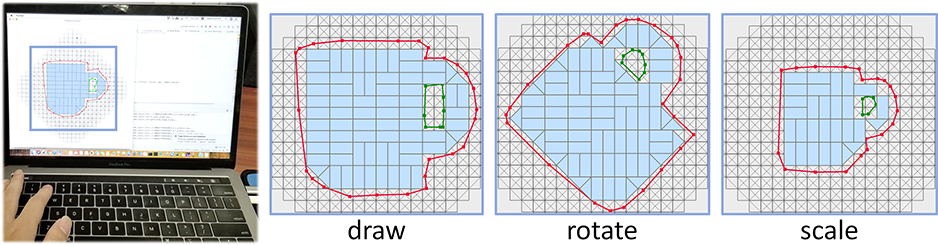}
	\vspace*{-1mm}
	\ifdefined\negativevspace
	\vspace*{-1.5mm}
	\fi
	\caption{We provide an interactive interface for tiling design, in which we can draw/load a shape, modify it, and preview the tiling solutions.}
	\label{fig:interface}
	\ifdefined\negativevspace
	\vspace*{-1mm}
	\fi
	\vspace*{-1.5mm}
\end{figure}
%%%%%%%%%%%%%%%%%%%%%%%%%%%%%%%%%%%%%%%%%%%%%%%%%%%%

%%%%%%%%%%%%%%%%%%%%%%%%%%%%%%%%%%%%%%%%%%%%%%%%%%%%
\begin{figure*}[t]
	\centering
	\includegraphics[width=0.99\linewidth]{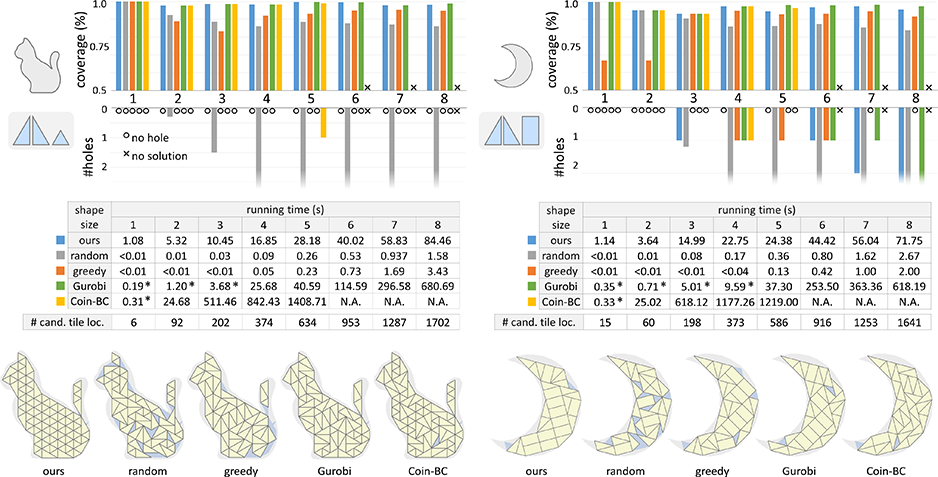}
	\ifdefined\negativevspace
	\vspace*{-1.5mm}
	\fi
	\caption{Comparing the performance (tiling coverage, number of holes, and running time) of our method with four alternatives on tiling the \textsc{Cat} and \textsc{Moon} shapes for eight different shape sizes.
	Note that ``\# cand. tile loc.'' denotes the number of candidate tile locations (solution space). 
	The results here show that our method consistently outperforms others (see Section~\ref{ssec:evaluations} for details), and its running time increases roughly linearly with the number of candidate tile locations.
	\final{On the bottom, we show tiling solutions produced by the five methods for \textsc{Cat} of shape size five (left) and \textsc{Moon} of shape size four (right).}
	}
	%\caption{Comparing the performance of our method with four alternatives (see the legend on bottom left) for tiling two different shapes at eight different sizes.}
	\label{fig:comparison}
	\ifdefined\negativevspace
	\vspace*{-1mm}
	\fi
\end{figure*}
%%%%%%%%%%%%%%%%%%%%%%%%%%%%%%%%%%%%%%%%%%%%%%%%%%%%

\ifdefined\negativevspace
\vspace*{-3pt}
\fi
\paragraph{Tiling generation.}
We compiled a set of 107 silhouette images as test shapes.
These images were collected by Internet image search and from two public data sets: the MPEG-7 data set~\cite{latecki2000shape} and animal shapes from~\cite{bai2009integrating}.
Here, we selected the shapes that (i) have distinctive shape features and (ii) do not possess delicate structures such as long and thin tails.

Then, we ran Algorithm~\ref{alg:algo_tiling} on a workstation with 16 CPUs and 125 GB memory to produce tilings on these test shapes.
Since our tiling problem has multiple objectives, we may not always avoid holes when putting more emphasis on maximizing the tiling coverage.
Thanks to the performance of TilinGNN, we are able to quickly run the tiling procedure multiple times (20 times, in practice), and take the tiling solution with a larger tiling coverage; if multiple results produce the same coverage area, we select the one with \ed{a} larger total length of contacting edge segments based on loss term $L_\text{e}$ (see Section~\ref{ssec:training_loss}).
Also, we can run multiple threads of the tiling procedure in parallel for better performance.
See again Table~\ref{fig:statistics_tileset} for the test time performance with different tile sets.

\subsection{Tiling solutions}
\label{ssec:results}

%\subsubsection{Visual results}

%%%%%%%%%%%%%%%%%%%%%%%%%%%%%%%%%%%%%%%%%

Figure~\ref{fig:gallery} presents a gallery, showcasing \phil{tiling} solutions produced on 36 different shapes by our learn-to-tile approach.
%using different tile sets.
%
These shapes have different sizes, convexity, and topologies.
%(some with holes).\phil{only one has hole... so, I removed this}
%
The average time for our approach to produce these tilings is only 23.5 seconds.
%See also supplementary material for more results.
\phil{Note that all tiling solutions presented in the paper are produced by Algorithm~\ref{alg:algo_tiling}, without any post refinement.}

%\paragraph{Statistics on tiling generation.}
%Figure~\ref{fig:gallery} shows a gallery of tiling of various shapes using different tile set.
%These results demonstrate that our method is able to generate tiling solutions for tiling regions of different shapes, sizes, topologies (with holes) and %convexity, 
%using different tile sets of different shapes and counts.
%Although the training data set are shapes in polar coordinates and are simple polygons (i.e., without holes inside),
%our network can generalize well to tile shapes that can not be represented in polar coordinates, and shape contains holes.
%The average running time for these shapes is only 23.5 seconds.
%See also supplementary material for more results.

%%%%%%%%%%%%%%%%%%%%%%%%%%%%%%%%%%%%%%%%%

\ifdefined\negativevspace
\vspace*{-3pt}
\fi
\paragraph{Tiling with different tile sets.}
Also, we employ tilinGNN models trained on different tile sets to tile the same shape, so as to explore the robustness of our method to variations in tile set.
Figure~\ref{fig:shape_with_diff_tiles} shows two sets of results on tiling the {\sc Dove} and {\sc Cat} shapes using seven different tile sets.
Our approach can generate interesting tiling solutions for the shapes and reproduced some featured regions using different tile arrangements; see particularly the wings of the different {\sc Dove}s and the tail of the different {\sc Cat}s for examples.

%\paragraph{Adapt to different tile sets.}
%To explore our method's robustness to variations in the tile set, we tile the same tiling region with six different tile sets.
%Figure~\ref{fig:shape_with_diff_tiles} shows the results.
%Our method can generate seamless tiling solutions for using different tiles, and to create featured regions using different tile arrangements, see the %wings of the \textsc{dove} and the tail of the \textsc{cat} for example.

%%%%%%%%%%%%%%%%%%%%%%%%%%%%%%%%%%%%%%%%%

\ifdefined\negativevspace
\vspace*{-3pt}
\fi
\paragraph{Interactive interface.}
Given a shape to tile, if we scale or modify its contour, the tiling solutions will likely change.
We provide an interactive interface for loading or drawing a shape and interactively transforming or editing the shape's contour; see Figure~\ref{fig:interface}.
%\phil{just to make this reference on the same page as the figure}
Since the contour is fixed on the tile set's superset by manual actions on the interface, we can skip step 1 in the tiling procedure (see Section~\ref{ssec:test}) and provide a preview of the tiling in one to three seconds, depending on the number of candidate tile locations in the contour.
Please see our supplemental video for a demonstration.

%\begin{itemize}
%	\item fast, can run on a mac book pro without GPU
%	\item user can use the interface to draw the shape, load and change shape boundary
%	\item with the interface, we do not need the top K cropping
%	\item 
%\end{itemize}

%%%%%%%%%%%%%%%%%%%%%%%%%%%%%%%%%%%%%%%%%

\ifdefined\negativevspace
\vspace*{-3pt}
\fi

\rz{
\paragraph{Mosaic-style tiling of large regions.}
To produce larger tilings, we can subdivide the domain to be tiled into moderately sized ``super-tiles'' and apply TilinGNN to tile each 
super-tile with exact conformation to its boundary. Then, by cropping the tiled region using the input image (i.e., selecting all tiles which
lie entirely inside the image boundary) and applying the colors 
from the image, we can produce mosaic-style tilings, as shown in Figure~\ref{fig:large_tilings}.}
%showcases several results, which contain XXX, XXX, and XXX tiles, respectively, for the results shown on XXX.
%\todo{work-in-progress

%%%%%%%%%%%%%%%%%%%%%%%%%%%%%%%%%%%%%%%%%

\paragraph{Note.}
Due to our problem formulation, the cropping process keeps only the candidate tile locations that are fully inside the given 2D shape (see Figure~\ref{fig:overview}(b)).
In fact, it is possible to slightly modify our implementation, such that we can relax this restriction and consider candidate tile locations that are {\em partially outside\/} the given 2D shape.
\rz{Such a modification may help to avoid holes and lead to less jagged boundaries in the generated tiling solutions.}

%All tiles in the tiling results are located fully inside the tiling region, due to our problem formulation. We can make small changes to the cropping procedure (see again Figure~\ref{fig:overview}(b)) to allow tiles to partially extrude out of the tiling region. While this change may help avoid larger holes, it may easily lead to jerky boundaries in the tiling solutions.}

%%%%%%%%%%%%%%%%%%%%%%%%%%%%%%%%%%%%%%%%%%%%%%%%%%%%%%%%%%%%%%%%%%%%%%%%%%%%%%%%%%

%\subsubsection{Quantitative comparison}
\subsection{Evaluations}
\label{ssec:evaluations}

%%%%%%%%%%%%%%%%%%%%%%%%%%%%%%%%%%%%%%%%%

\paragraph{Comparison with alternative approaches.}
We compared our method with four alternatives.
%, by which we created tiling solutions, specifically for maximal tiling coverage and minimal holes, without tile overlaps.
%
The first two are
(i) a random search framework that uses
%the tiling procedure in 
Algorithm~\ref{alg:algo_tiling} but replaces the TilinGNN output probabilities with random values in $[0,1]$; and
(ii) a greedy strategy that follows existing shape packing methods (e.g.,~\cite{Kwan-2016-2DCollage}) to iteratively select the tile that shares the longest boundary with the current partial tiling solution.
Besides, we employed two state-of-the-art integer programming solvers:
\final{(iii) a specialized integer programming solver, Gurobi~\shortcite{gurobi}, which has been shown to perform fast in many combinatorial optimization problems~\cite{luo-2015-legolization, peng-2019-checkerboard, Wu2018-stitchmeshing}; and}
%(iii) an open-source constraint-based solver, Gecode~\cite{gecode}; and 
(iv) a \final{branch \& cut} mixed-integer solver (Coin-BC)~\shortcite{coin-bc}.
%, which is good at linear integer programming.
%
To employ them, we formulate our problem using a constraint modeling language, Minizinc~\cite{csplib-minizinc}, \final{in which} we model the selection of each node as a binary variable, set the tile overlaps as hard constraints, and define an objective function (similar to $L_a$ and $L_e$ in Section~\ref{ssec:training_loss} but without logarithms) to maximize the tiling coverage and total length of shared edge segments. \final{See supplementary material part B for the code.}

Figure~\ref{fig:comparison} reports the comparison results on tiling two different shapes: 
the percentage of tiling coverage and number of holes (top), running times (middle), \final{and visual comparisons (bottom)}, where we tile the input shapes for eight different sizes (20\% to 90\% of the tile set's superset).
For each result (5 methods $\times$ 2 shapes $\times$ 8 sizes), we run the method ten times and take the average.
Also, to sense the size of the search space, we provide the number of candidate tile locations per case on the bottom row of the figure tables.
%of Figure~\ref{fig:comparison}.
%

The random strategy, in fact, serves as a control for revealing if TilinGNN is helpful to produce better tilings.
Comparing random's (grey) with ours (blue) in Figure~\ref{fig:comparison}, we can see that our approach produces tilings consistently with larger coverage and fewer (or no) holes.
Hence, the probability outputs from TilinGNN help improve the tiling solutions.
On the other hand, the greedy strategy (orange) is fast and often produce\final{s} reasonable solutions.
%able to produce good-quality \xh{acceptable?} solutions.
However, it relies on a local heuristic, so it cannot produce high tiling coverage\final{.} %, compared with ours.

For \final{Gurobi} (green) and Coin-BC (yellow), they are designed for solving discrete combinatorial optimization problems, essentially search problems, meaning that they have to iteratively explore and prune the solution space using strategies such as tree search pruning and \final{the} cutting plane method.
\final{In practice, they progressively output feasible solutions of high objective values, and do not stop until they find the optimal solutions (after exploring the entire search space with certain pruning).
Hence, for a fair comparison with them, we can either {\em restrict their running times to be the same as ours} then compare the quality of the tiling solutions, or {\em stop them when they produce tiling solutions of similar/better quality (coverage) as ours} then we can compare the running times.
For the former strategy, we found that these solvers are only capable of producing tiling solutions for smaller problems (marked with asterisks in the figure tables).
Therefore, we took the latter strategy and stop these solvers based on quality.
However, for Coin-BC, we found that for some larger problems, it could not produce tiling solutions even running for a long time, so we stop it if it runs 50 times more than our method (marked by N.A. in the figure tables).}
%
%
% Gurobi - first two scales
% Coin-BC - first one scale
%
\final{Overall, from the results shown in Figure~\ref{fig:comparison}, we can see that Gurobi and Coin-BC can produce plausible solutions, but they require much longer running time than ours, especially for larger problems.
% the number of decision variables equals the number of candidate tile locations, which increases exponentially with the shape size (see again Figure~\ref{fig:comparison}).
% Hence, the two solvers could not find tiling solutions in reasonable time, comparable to our approach, when the shape is \final{large}.
As for our method, its running time increases roughly linearly with the number of candidate tile locations, as shown in Figure~\ref{fig:comparison}.}

\begin{figure}[t]
	\centering
	\includegraphics[width=0.99\linewidth]{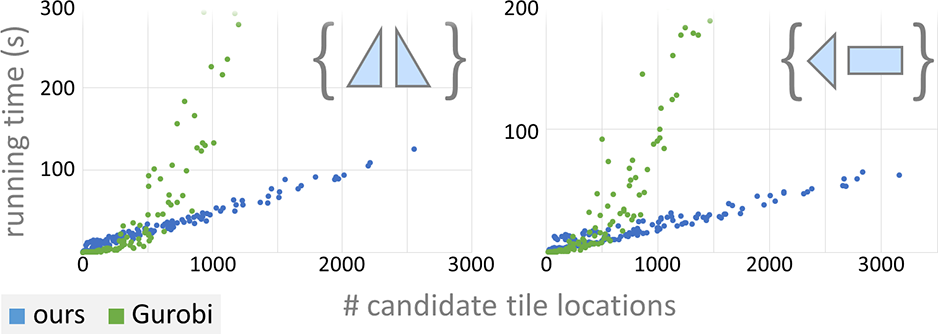}
	\vspace*{-0.5mm}
	\ifdefined\negativevspace
	\vspace*{-1.5mm}
	\fi
	\caption{Linearity test.
		These scatterplots show the running time of our method (blue) \final{and Gurobi (green)} for tiling 20 shapes of varying sizes. 
		%\final{To clearly show the trend of the two methods, we ignore the points with running time above 600s (left) and 300s (right), respectively.}
		The plots reveal the running-time linearity of our method with respect to the number of candidate tile locations\final{, while Gurobi grows much faster.}}
%	\caption{Scalability test.
%		Top: plotting of the running time increasing with the number of tile placements.
%		Bottom: area of the selected tiles increasing with the area of the tiling region.}
	\label{fig:scalability}
	\ifdefined\negativevspace
	\vspace*{-1mm}
	\fi
	\vspace*{-0.5mm}
\end{figure}

\ifdefined\negativevspace
\vspace*{-3pt}
\fi
\paragraph{\final{Running time analysis.}}
Next, we explore the running time of our method \final{and also the Gurobi solver} by using \final{them} to solve 320 tiling problems: 20 shapes $\times$ 8 sizes $\times$ 2 tile sets.
Here, we measured the time taken by \final{each method} to generate each result and recorded the associated numbers of candidate tile locations.
\final{Same as the previous experiment, we stop Gurobi from running if it produces tiling solutions of similar or better quality as ours.}
The \final{two} scatterplots (for the two different tile sets) shown in Figure~\ref{fig:scalability} report all the numbers, revealing that the time taken by our method increases roughly linearly with the number of candidate tile locations\final{, while the running time of Gurobi grows much faster.}
\final{For our method, t}he small fluctuations between results (sample points in the plots) of similar number of candidate tile locations are due to the number of rounds that the tiling procedure took.
%to produce the results.
The number of rounds is usually one or two, but having more than that would suddenly increase the overall running time.
\final{For Gurobi, while it solves small-scale problems (with less than $500$ candidate tile locations) quickly, it does not scale well for larger problems.}

\begin{figure}[t]
	\centering
	\includegraphics[width=0.99\linewidth]{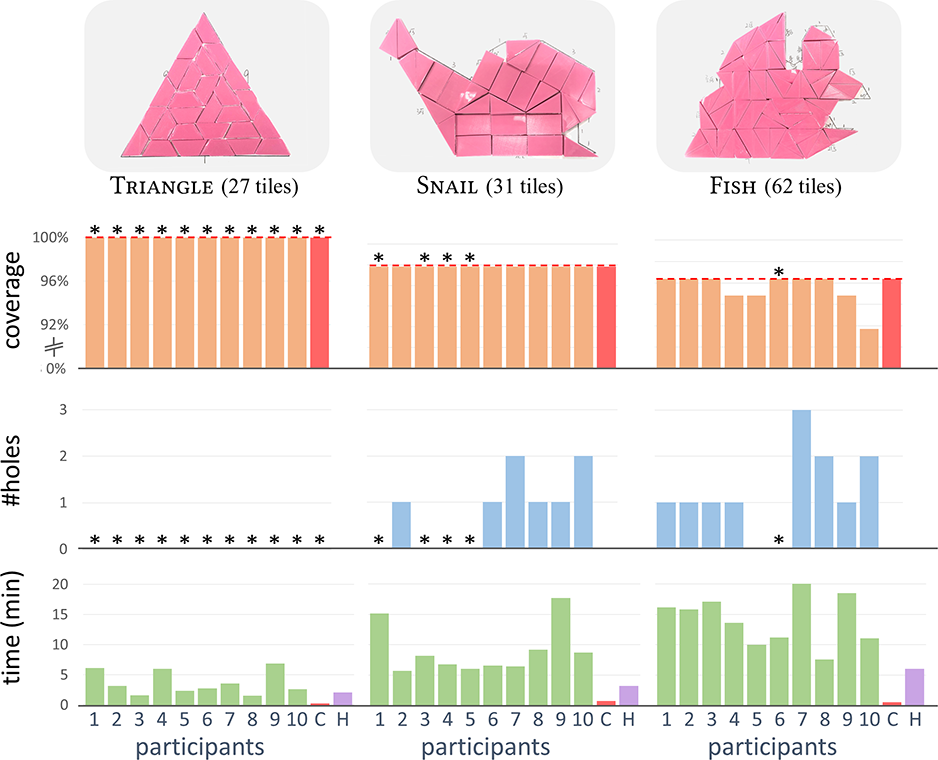}
	\vspace*{-1mm}
	\ifdefined\negativevspace
	\vspace*{-1mm}
	\fi
	\caption{Exploring human performance.
	We recruited 10 participants to solve three tiling problems (top) and measured their performance in terms of tiling coverage percentage, number of holes, and time spent.
	Note that the red bars show the performance of our method (denoted by C), whereas the violet bars show the times taken to follow our method's solutions to manually create the tilings by hands (denoted by H).
	Also, the asterisks mark the participant solutions that attain the same quality as our method's.}
	%of 27, 31, and 63 tiles 
	\label{fig:human_performance}
	\ifdefined\negativevspace
	\vspace*{-1mm}
	\fi
	\vspace*{-0.5mm}
\end{figure}

\ifdefined\negativevspace
\vspace*{-3pt}
\fi
\paragraph{Human performance.}
Next, to explore how humans solve the tiling problems faced by TilinGNN, we picked three tiling solutions produced by TilinGNN with increasing difficulty levels as the tiling problems in this experiment; see Figure~\ref{fig:human_performance} (top).
%, using different tile sets to fill different shapes.
%
For each problem, we 3D-printed the tiles of the tile set, computed the boundary contour of all the candidate tile locations, and further printed out the boundary contour on paper at a scale compatible with the 3D-printed tiles.
Hence, the participants could directly manipulate the 3D-printed tiles on the printed paper to work out their solutions.

Overall, we recruited 10 participants who are university students: 3 females \& 7 males, aged 22 to 26.
%, on a volunteering basis.
%They , and among them, \todo{XXX} had experience in building Tangram.
%
When the experiment started, we first introduced the goal of tiling to the participants, i.e., to tile a shape's interior with maximum area coverage and minimum holes, while not exceeding the shape's boundary.
Hence, they should try to arrange as many tiles as possible in their solutions.
Also, we showed them examples of holes, the 2D dimensions of the tiles, and \final{marked} the length of each side along the boundary contours \final{on paper}.
After that, each participant was given at most half an hour to work out their solutions for each tiling problem, in which we recorded the resulting tiling coverage, number of holes, and time spent.

Figure~\ref{fig:human_performance} reports the recorded quantities for all \final{the} participants and also for our method.
The left column shows the simplest case of tiling the {\sc Triangle} shape, where all participants found the optimal solution with 100\% coverage without holes and tile overlaps.
The other columns show two harder cases.
For {\sc Snail}, four participants found solutions of same quality as ours: 97.8\% coverage and without holes, whereas for {\sc Fish}, only one participant found a solution of same quality as ours: 96.3\% coverage and without holes.
For fair comparison on time, we measured the time taken to follow our method's solutions and manually \final{reproduce}
%create 
the tilings by two hands.
On average, the participants took 3.66 mins for {\sc Triangle}, 9.04 mins for {\sc Snail}, and 14.13 mins for {\sc Fish}, 
while using our method, we took only
0.24 + 1.82 mins for {\sc Triangle},
0.70 + 2.51 mins for {\sc Snail},
and 0.53 + 5.50 mins for {\sc Fish}.
The first numbers are our method's running times and second numbers are manual tiling times.

\begin{figure}[t]
	\centering
	\includegraphics[width=0.94\linewidth]{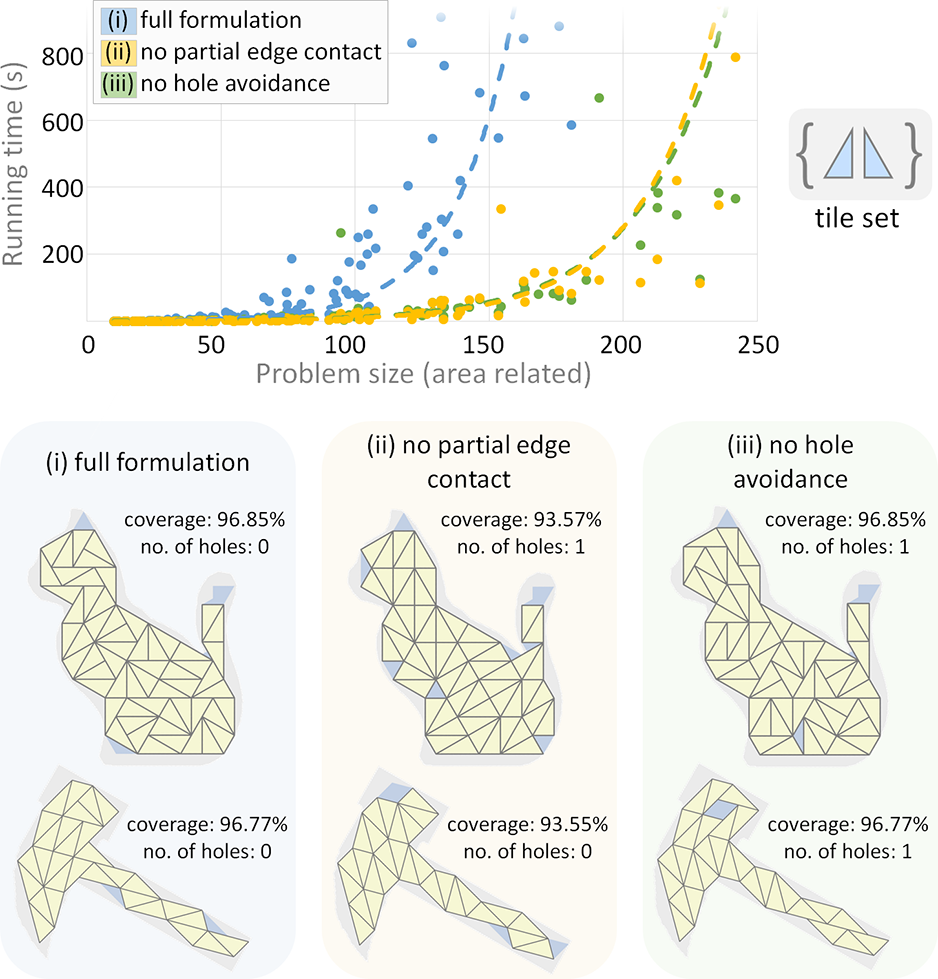}
	\vspace*{-1.5mm}
	\caption{\final{Exploring the computational complexity of considering partial edge contacts between tiles (setting (ii)) and explicit hole avoidance (setting (iii)).
	Top: scatterplots show the relationship between the running time and problem size for various settings, including the full formulation (setting (i)); comparing the plots for settings (ii) and (iii) against setting (i), we can see that our full formulation is more tedious to compute with, for not so small problem size, thereby revealing the complexity of partial edge contacts and explicit hole avoidance.
	Bottom: some of the associated tiling solutions.}}
	%, showing that without partial edge contacts, T-junctions disappear between adjacent tiles, without explicit hole avoidance, holes may appear inside the tiling solutions.}}
	%
%	\caption{\final{Comparing the computational complexity of considering 1) partial edge contacts between adjacent tiles and 2) explicit hole avoidance. The scatterplots (top) show that considering the two aspects increases the running time for the same problem size (x-axis), which is the area of the cropped superset in units of one triangle tile's area (top-right). We also show some of the tilings solutions in bottom to compare their coverage and number of holes.}}
	\label{fig:hole_junction} 
	\vspace*{-1mm}
\end{figure}

%GOAL:
%In the rebuttal you mention "Particularly, our method allows adjacent tiles to contact by a partial or full edge segment, so the tile connectivities are very different. Further, our method needs to explicitly avoid holes, thus making the computation more tedious.". This is something you need to demonstrate as part of the comparison to IP methods.

\vspace*{-3pt}
\final{
\paragraph{Complexity: partial edge contacts and hole avoidance.}
To explore the computational complexity of considering {\em partial edge contacts\/} (T-junctions) between adjacent tiles and {\em explicit hole avoidance\/} in \phil{our} formulation, we conducted an experiment with the following three different settings:
%(which are absence in previous works such as~\cite{peng-2019-checkerboard})
%
(i) our full formulation;
(ii) our full formulation without candidate tile locations connected with partial edge contacts; and
(iii) our full formulation without an $L_e$ term for explicitly avoiding holes in the tiling solutions.
In detail, we employed the 20 shapes of 8 different sizes as before, and used Gurobi~\shortcite{gurobi} to search for tiling solutions in each case and setting.
Note that we employed Gurobi in this experiment because it can explore the entire search space and find optimal solutions, given sufficient time.
By this means, we can sense the complexity of different settings.

Figure~\ref{fig:hole_junction} (top) presents scatterplots that reveal the relationship between running time and problem size for each setting, whereas Figure~\ref{fig:hole_junction} (bottom) presents some of the associated tiling solutions.
Note, for a fair comparison between settings with and without partial edge contacts, we do not take the number of candidate tile locations as the problem size but use the area covered by the cropped superset, where a single triangle tile is said to have one unit area (see Figure~\ref{fig:hole_junction} (top) for the tile set).
From the plots shown on top of the figure, we can see that if we do not consider partial edge contacts (setting (ii)) or do not explicitly avoid holes (setting (iii)), the running time would decrease significantly for not-so-small problems of same problem size (x-axis), meaning that if we consider partial edge contacts (settings (i) vs. (ii)) or explicitly avoid holes (settings (i) vs. (iii)), Gurobi would require more time in the search and the computation would become more tedious.
Also, we use the exponential trendline tool in Excel to fit the dashed lines shown in the plots for each of the three settings to further reveal the relationship between running time and problem size in each setting.}

\begin{table}[t]
	\centering
	\caption{Network and loss analysis.
	Performance (tiling coverage \%) of TilinGNN re-trained for various situations of ablating or modifying the network architecture and loss function.
	Comparing with the re-trained models, the full method (bottom right) consistently achieves better performance.}
	\ifdefined\negativevspace
	\vspace*{-1mm}
	\fi
	\includegraphics[width=0.99\linewidth]{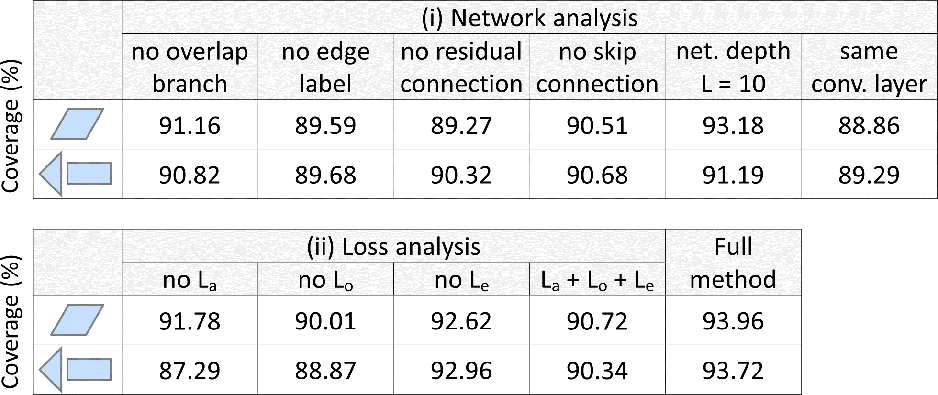}
	%\caption{Network effective of terms in losses, the architecture and the training of our network.}
	\label{fig:net_analysis}
	\ifdefined\negativevspace
	\vspace*{-1mm}
	\fi
\end{table}
%%%%%%%%%%%%%%%%%%%%%%%%%%%%%%%%%%%%%%%%%%%%%%%%%%%%

\subsection{Network Analysis}
\label{ssec:net_analyze}
In the following, we present analysis on various aspects of our method by re-training TilinGNN on two different tile sets and applying the re-trained models to tile 20 test shapes, each having three different sizes, i.e., $30\%$, $50\%$, and $80\%$ of a tile set's superset.

%The test shapes have different complexities, topologies, and aspect ratio; see the supplementary material \todo{}.

%, see the three tile sets in Figure~\ref{fig:statistics_tileset} that are marked *.
%
%We train each modified network or with modified losses with the same number of epochs with our method.
%We compare the effective of each model by evaluating it on 20 different shapes.
%The selected shapes have different complexity, symmetry property, convexity, and aspect ratio, see supplementary material for these shapes.
%For each shape, we crop $\mathcal{P}$ using three scales $0.3$, $0.5$, $0.8$ relatively large with $\mathcal{P}$ to synthesis adjacency graphs with %different sizes.
%
%After training, we report the two numbers:
%\begin{itemize}
%	\item the average coverage of the tiling solutions on the tile placements.
%	\item the average number of holes in the generated tiling solutions.
%\end{itemize}

%%%%%%%%%%%%%%%%%%%%%%%%%%%%%%%%%%%%%%%%%

\ifdefined\negativevspace
\vspace*{-3pt}
\fi
\paragraph{(i) Network architecture analysis.}
To evaluate the architecture design of TilinGNN, we removed from it the following network components, one at a time, and re-trained new TilinGNN models:
%, and re-tested the models with the test shapes:
%
\begin{itemize}
\item no overlap branch, by removing all overlap aggregation modules and all element-wise products at the end of each feature aggregation layer (see Figure~\ref{fig:net_architec});
\item no edge labels, by setting all elements in $\{ \mathbf{e}_j \}$ to one;
\item no residual connections, by removing the residual connection in both \final{the} neighbor and overlap aggregation modules (see Section~\ref{ssec:network_arch} and Figure~\ref{fig:graph_conv});
\item no skip connections, by interpreting only the latent feature of the final layer, i.e., by taking $\mathbf{F}^L$ as $\mathcal{F}$ without $\mathbf{F}^0$ to $\mathbf{F}^{L-1}$;
\item setting network depth $L = 10$, which is the number of feature aggregation layers (see Figure~\ref{fig:net_architec}); and
\item using the edge-conditioned convolution layer \final{not only in the neighbor aggregation module but also in the overlap aggregation module (see Section~\ref{ssec:network_arch}) by using all-one vectors as the input features and replacing Eq.(\ref{eq:ovl_conv}) with Eq.(\ref{eq:nbr_conv}) in the overlap aggregation module.}
% setting the input edge vectors for overlap edges as all-one vectors.
%in both training and testing.
%\phil{what does it mean by ``by setting the input edge vectors for overlap edges as all-one vectors''? Does it give the effect that the overlap aggregation module uses the edge-conditioned convolution layer? XH?}
%\xh{Because the edge-conditioned convolution must consume edge features.}
%
\end{itemize}

%%%%%%%%%%%%%%%%%%%%%%%%%%%%%%%%%%%%%%%%%
\ifdefined\negativevspace
\vspace*{-3pt}
\fi
\paragraph{(ii) Loss analysis.}
Besides, we re-trained TilinGNN models for the following \final{four} situations to analyze the loss function design:
\begin{itemize}
\item removing $L_\text{a}$, $L_\text{o}$, or $L_\text{e}$ from the overall loss; and
\item summing the three loss terms ($L_\text{a} + L_\text{o} + L_\text{e}$) rather than multiplying them as the overall loss.
\end{itemize}

%%%%%%%%%%%%%%%%%%%%%%%%%%%%%%%%%%%%%%%%%

For comparison purpose, we compute the percentage coverage of each tiling solution, and consider the average percentage over the test shapes for each re-trained TilinGNN model as its performance.
Table~\ref{fig:net_analysis} reports the performance for all the above ten cases, including also the performance of our full method.
Comparing the numbers shown in the table, we can see that if we ablate TilinGNN by removing the overlap branch, by removing the edge labels, etc., or \phil{by} modifying its architecture or loss function, the tiling performance drops.
This reveals the contributions of individual network components and the loss function design to the full method.

\begin{figure}[!t]
	\centering
	\includegraphics[width=0.99\linewidth]{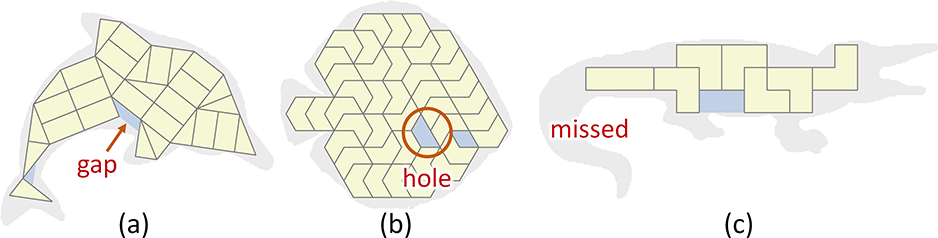}
	\ifdefined\negativevspace
	\vspace*{-1.5mm}
	\fi
	\caption{Potential issues in our tiling solutions.}
	\label{fig:fail_cases}
\end{figure}
%%%%%%%%%%%%%%%%%%%%%%%%%%%%%%%%%%%%%%%%%%%%%%%%%%%%
\begin{figure}[t]
	\centering
	\includegraphics[width=0.98\linewidth]{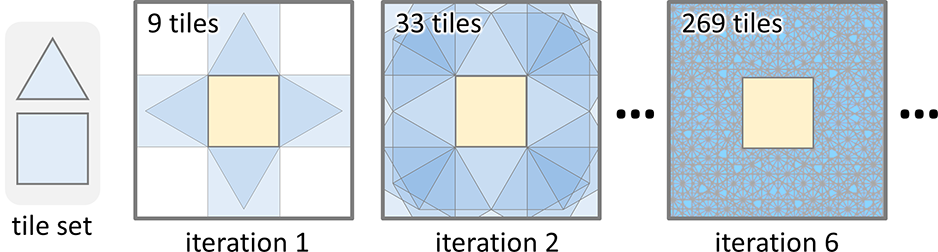}
	\caption{\finaltwo{The example square-triangle tile set~\cite{wiki:aperiodic_tileset} does not induce a periodic grid for the candidate tile locations. % in the gray box.
	In the recursive tile enumeration, most newly generated candidate tiles do not coincide with the existing ones, leading to an unbounded %exponential growth 
expansion of candidate tiles.}}
	%\caption{\xh{Example tile set that produces infinite candidate tile placements inside the gray border. In our recursive tile enumeration process, most newly generated candidate tiles do not coincide with existing ones, making an explosive amount of candidate tiles.}}
	\label{fig:irregular}
\end{figure}

\section{Conclusion}
%\paragraph{limitation and future works}
%\begin{itemize}
%	\item train with reinforcement learning.
%	\item 3D tiling.
%   \item automatically select template shapes
%   \item the scalability of the problem depends on the size of the candidate graph
%   \item our results can not gurantee symmetry.
%	\item can not deal with complex tile set.
%   \item long training time
%   \item limited by GPU memory
%   \item not continuous alignment
%\end{itemize}

We presented the first deep-learning-based approach to generate non-periodic tilings on given 2D shapes.
% by formulating the problem over a graph and 
%, so that we can adopt a graph neural network to work on the problem and help generate tiling by 
%adopting a graph neural network to predict tile placements.
%
Overall, our approach has three main contributions.
First, we model tiling as a graph problem with nodes and edges representing the candidate tile locations and tile connectivity, so we can adopt a graph neural network to solve the problem by learning features to predict probabilities for tile placements.
Second, we design TilinGNN, a two-branch graph convolutional neural network model with the neighbor and overlap aggregation modules to progressively aggregate features along graph nodes for producing tilings with maximized coverage, while avoiding holes and tile overlaps.
Third, we formulate these criteria as loss terms defined on the network outputs, so we do not require manual preparation of ground-truth tilings, and the network training can be self-supervised.
\finaltwo{Our method works on many different forms of tile sets} and allows contacts at partial segments.
%, which are absent in typical tiling solutions.}
%\TODO{Do we need to keep this T-junction claim?}

We \phil{performed} various experiments to evaluate the quality of our approach, and \phil{presented} a variety of tiling results to demonstrate the possibility of using a neural network to learn to produce tilings.
\final{Experimentally}, the running time of our network is roughly linear to the number of candidate tile locations, which far exceeds the performance of conventional combinatorial search tools.

%Also, we demonstrated the possibility of using a neural model to solve many challenging tiling problems.

%%%%%%%%%%%%%%%%%%%%%%%%%%%%%%%%%%%%%%%%%%%%%%%%%%%%%%%%

\ifdefined\negativevspace
\vspace*{-3pt}
\fi

\paragraph{Limitations.}
As a first attempt to generate tilings by machine learning, our method still has several limitations.
\finaltwo{First, as explained earlier in Section~\ref{ssec:tile_set}, TilinGNN is a selection network that learns to select from a finite set of candidate tile locations, so we consider tile sets that the induced candidate tile locations form periodic grids, e.g., see the superset in Figure~\ref{fig:overview}(a).
Hence, TilinGNN cannot deal with all kinds of tile sets, e.g., the Penrose tile set and the square-triangle tile set shown in Figure~\ref{fig:irregular}.}
%\phil{TODO: Because our network learns to select from the set of {\em all} candidate tile placements, thus the set should neither be infinite nor incomplete.}
%
%for our network to work, we have to be able to enumerate all the candidate tile placements. Therefore, our network cannot handle tile sets that induces infinite number of candidate tile placements, e.g., the square and equilateral triangle tiles shown in Figure~\ref{fig:irregular} and the Penrose tilings.}
%\xh{Also,} the tiling solutions that can be produced by TilinGNN are limited by the choice of the 
%candidate tile sets. While in general, the tilings obtained are non-periodic, one can still observe
%many patches formed by regular tessellations from results shown in the paper\xh{, which is inherited from the regularities that existed in our current choice of  the candidate tile sets (see Section~\ref{ssec:tile_set}).}
%(see the inset figure).
% However, there is no refinement being modeled or enforced by our network, while the network does not explicitly enforce irregularity either. 
%The tiling regularities in the results were inherited from the regularities  that existed in our current choice of candidate tile sets (see Section~\ref{ssec:tile_set}). And as such, the current TilinGNN is unable to produce aperiodic pattern such as Penrose tilings.
%The network training is hard to converge, due to the huge solution space and diverse solution patterns.
%
\finaltwo{Also, it} is unclear how the network can effectively enforce symmetries in the tiling. We did attempt to include 
symmetry as a loss term, but network training became an issue, where the loss did not converge.
In general, compared to existing IP solvers such as Gurobi, our method lacks the ability to impose
hard constraints or tweak the objective functions.

Furthermore, results from TilinGNN may not always avoid gaps (see Figure~\ref{fig:fail_cases} (a))
%due to the cropping location, 
and holes (see Figure~\ref{fig:fail_cases} (b)), due to the multi-objective goal, which also needs to 
address the tiling coverage and overlap avoidance.
Also, the tiling results may miss shape structures that are typically thin (see Figure~\ref{fig:fail_cases} (c)), 
if the shape does not fully cover surrounding candidate tile locations in the cropping.
%
%As 
\phil{At} last, from a computational standpoint, the scalability of our current implementation, or the 
maximum problem size that can be solved, is limited by the available GPU memory.

%\setlength{\columnsep}{2.8mm} % wrapfigure margin
%\begin{wrapfigure}{r}{0.35\columnwidth}
%	\vspace{-10pt}
	%\centering
%	\includegraphics[width=0.30\columnwidth]{images/triangle-square_irregular.png}
%	\hspace{-30pt}
%	\vspace{-20pt}
%\end{wrapfigure}
%

%\begin{itemize}
%	\item cannot always avoid holes or gaps on the boundary, because we optimize it as an objective, not hard constraint.
%	\item can not deal with over-thin structure.
%	\item symmetry
%	\item can not support arbitrary tile set, e.g., Penrose tilng, since ...
%\end{itemize}

%%%%%%%%%%%%%%%%%%%%%%%%%%%%%%%%%%%%%%%%%%%%%%%%%%%%%%%%
\ifdefined\negativevspace
\vspace*{-3pt}
\fi
\paragraph{Discussion.}
Combinatorial optimization is often a challenging but necessary problem to face in computer graphics research, whenever we have discrete options to choose in our problem.
In this work, we present a novel approach for tiling generation, where we demonstrate that discrete options, together with their constraints, can be formulated as a graph problem.
Consequently, we can then adopt and formulate a graph neural network to work on the problem.
Our overall framework is general and has great potential for solving other combinatorial problems.
Having said that, if we can connect the discrete options into a graph structure and model the constraints in the graph, we can then design features to be produced and aggregated at the nodes and along the edges for graph learning.
%neural network can possibly be adopted to work on the problem.
%
Taking the hybrid mesh problem~\cite{Peng-2018-Hybrid-Meshes} as an example, to decide whether to use a triangle or a quad at any local area is a discrete option, and such a local option affects the other local options that surround it.
Hence, if we can model the relations between local options into a graph, a graph learning model may be adopted to learn to create such mesh.
Besides, computational assembly problems~\cite{luo-2012-Chopper,masonry-2014-Assembly,xu-2019-technic-lego} often involve discrete options, in which we are strongly interested in exploring machine learning approaches to solving these problems.

%one of the key point we should discuss in the last section is how our work may be more general than just tiling. We try to stimulate the idea and show %potential of its generality in solving other hard problems. 
%
%What can others learn from us:
%\begin{itemize}
%	\item use a graph structure to encode discrete and irregular problem instances.
%	\item how to generalize the continuous neural network setting to solve discrete problems.s
%\end{itemize}

%%%%%%%%%%%%%%%%%%%%%%%%%%%%%%%%%%%%%%%%%%%%%%%%%%%%
\begin{figure*}[t]
	\centering
	\includegraphics[width=0.99\linewidth]{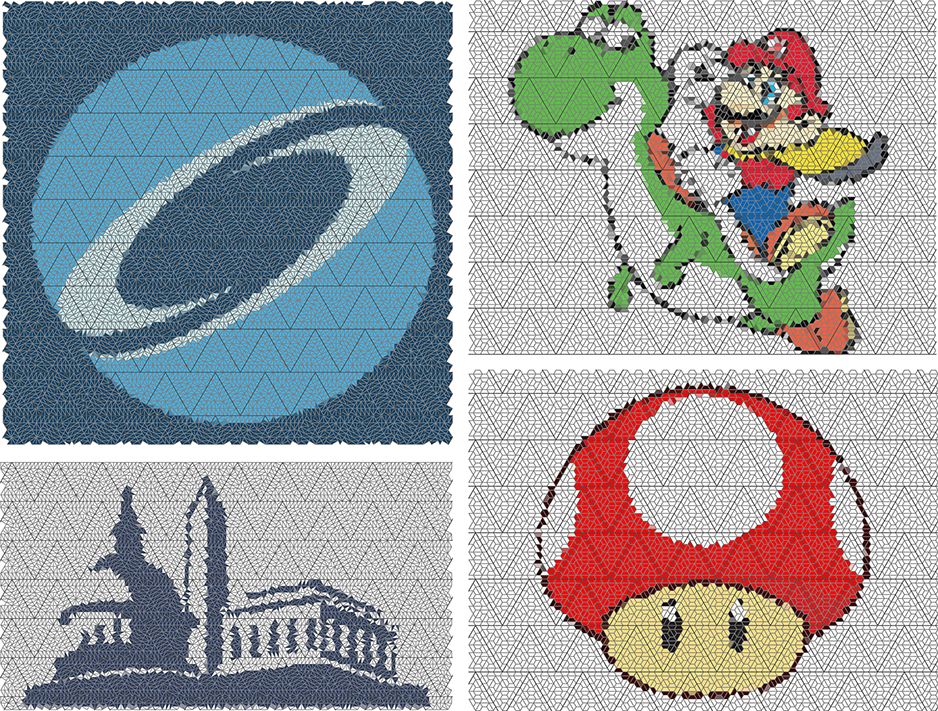}
	\ifdefined\negativevspace
	\vspace*{-1mm}
	\fi
	\caption{\rz{Large} mosaic-style tilings produced by subdividing \rz{the 2D plane into triangular super-tiles (shown with slightly darker boundaries)\/}, employing TilinGNN to tile each \rz{super-tile}, \rz{selecting all tiles lying entirely inside the input image, and} assigning a single color from the image to each generated tile.}
	\label{fig:large_tilings}
	%\vspace*{-1mm}
\end{figure*}
%%%%%%%%%%%%%%%%%%%%%%%%%%%%%%%%%%%%%%%%%%%%%%%%%%%%

\begin{acks}
\final{We thank all the anonymous reviewers for their comments and feedback. 
Figures 2(a), (b), and (c) are courtesy of National Trust of Australia (Victoria), 
Erhan Cubukcuoglu, and Katie Walker, respectively. This work is supported in part 
by grants from the Research Grants Council of the Hong Kong Special Administrative 
Region (Project no. CUHK 14201717 and 14201918), NSERC grants (No. 611370), and 
gift funds from Adobe.}
\end{acks}

% Bibliography
\bibliographystyle{ACM-Reference-Format}
\bibliography{learntotile}

\end{document}